%% file: iclr2026_conference.tex
\documentclass{article} % For LaTeX2e

\PassOptionsToPackage{dvipsnames}{xcolor}
\usepackage{iclr2026_conference,times}

\input{math_commands.tex}

\usepackage{hyperref}
\usepackage{url}
\usepackage{graphicx}
\usepackage{amsmath}
\usepackage{amssymb}
\usepackage{comment}
\usepackage{listings}
\usepackage{fontawesome} 
\usepackage{booktabs}
\usepackage{nicematrix}
\usepackage{arydshln}
\usepackage{balance}
\usepackage{wrapfig}

\definecolor{revision}{rgb}{0, 0, 0}
\arxivcopy

\title{\includegraphics[width=1em]{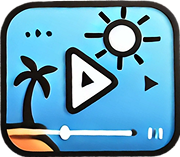} LLaVAction: evaluating and training multi-modal large language models for action understanding}

\begin{document}

\author{
    Haozhe Qi$^*$, Shaokai Ye$^*$, Alexander Mathis$^{**}$\small\faEnvelope, Mackenzie Weygandt Mathis$^{**}$\small\faEnvelope \\
    École Polytechnique Fédérale de Lausanne (EPFL), Lausanne \\
    {{\tt\small \faEnvelope \space mackenzie.mathis@epfl.ch, alexander.mathis@epfl.ch} \small
    $^*$Co-first, $^{**}$Co-senior}
}

\maketitle

\begin{abstract}
Understanding human behavior requires measuring behavioral actions. Due to its complexity, behavior is best mapped onto a rich, semantic structure such as language. Emerging multimodal large language models (MLLMs) are promising candidates, but their fine-grained action understanding ability has not been fully examined. In this work, we reformulate EPIC-KITCHENS-100, one of the largest and most challenging egocentric action recognition datasets, into a MLLM benchmark (EPIC-KITCHENS-100-MQA). We demonstrate that when we sample difficult answers based on specialist models as distractors, leading MLLMs struggle to recognize the correct actions. How can we increase the performance of MLLMs? We curated a supervised finetuning dataset that includes `hard' action recognition, temporal detection, captioning, and free-form question answering to improve models' diverse action understanding capabilities. We introduce a new model called LLaVAction that adds an action token to boost models' attention on visual tokens and a two-stage pipeline to obtain structured actions. LLaVAction greatly improves the MLLMs' ability of action understanding, achieving strong improvements on both MLLM benchmarks (21 points in accuracy over GPT-4o on EPIC-KITCHENS-100-MQA) and established action recognition benchmarks, suggesting that our methods prepare MLLMs to be a promising path forward for complex action tasks. 
Code, data, the benchmark, and models are available at \href{https://github.com/AdaptiveMotorControlLab/LLaVAction}{https://github.com/AdaptiveMotorControlLab/LLaVAction}.
% {\color{revision}{Code, data, benchmark and models will be available upon acceptance.}}
\end{abstract}

\section{Introduction}
\label{sec:intro}

Understanding human behavior is a complex challenge requiring multiple skills such as visual perception, knowledge about the world and reasoning capabilities. Current State-of-the-Art (SOTA) methods in action understanding tasks~\citep{chalk2024tim, liu2025opentad, shi2023temporal} typically rely on visual foundation models to imbue those kind of priors~\citep{radford2021learning, wang2022internvideo, wang2023videomae}. However, they rely heavily on dataset-specific target heads and have limited language understanding ability, constraining their performance and especially generalizability. Recently, Multi-modal Large Language Models (MLLMs)~\citep{zhang2024video, li2024llava, openai2024gpt4ocard, wang2024qwen2} have shown great potential for learning language priors to help understand visual content, making them promising alternatives. 

\begin{figure}[t]
    \centering
    \includegraphics[width=.95\linewidth]{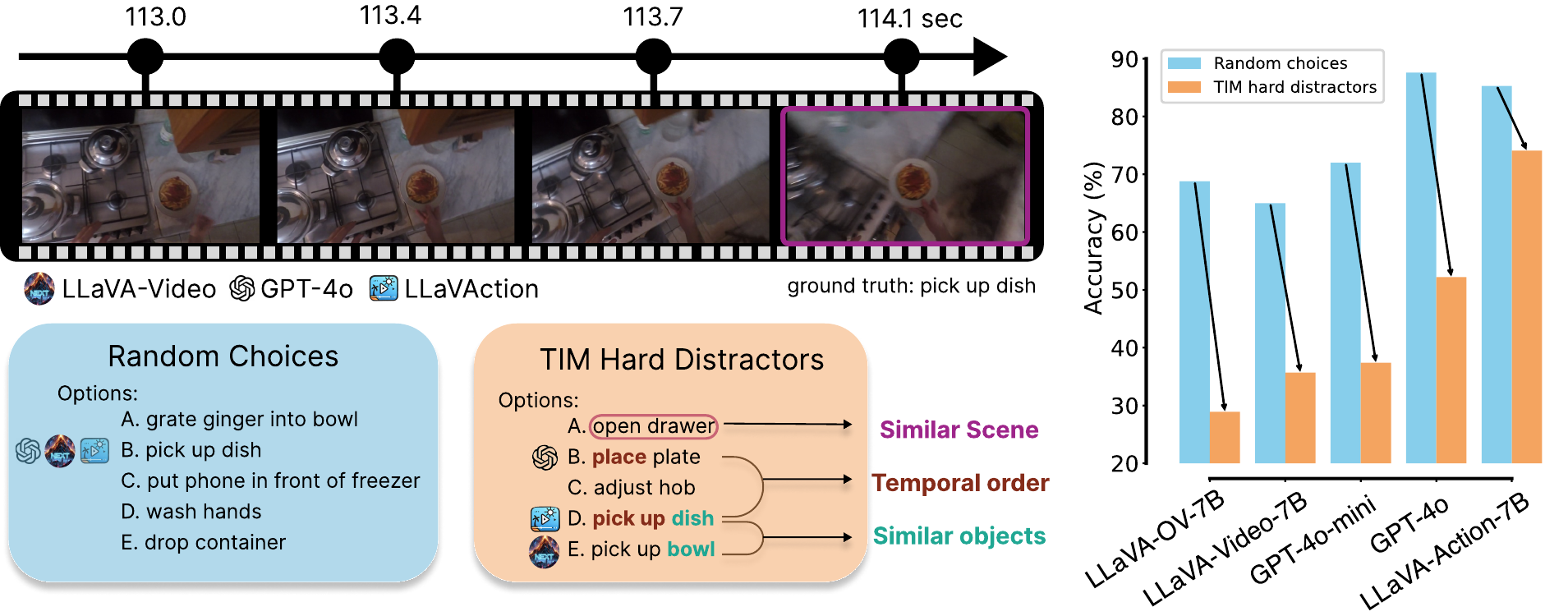}
    \vspace{-.2cm}
    \caption{\textbf{LLaVAction-7B.} Left: Qualitative inspection of distractors. We show an example clip with labels from random choices (which empirically is easy to solve), vs. our proposed harder benchmark with action labels generated by a SOTA specialist (TIM~\citep{chalk2024tim}). Our hard example mining strategy can automatically explore challenges such as temporal order and similar objects that are curated in other benchmarks. Right: While GPT-4o is strong when identifying correct answers among few random choices due to the large number of possible actions, it suffers in the harder benchmarking regime, and our method, LLaVAction outperforms GPT-4o.}
\vspace{-.5cm}
\label{fig:ek100-mqa}
\end{figure}

MLLMs take visual and text information as inputs and can directly output text. For training and evaluating MLLMs on action understanding tasks, existing datasets~\citep{kay2017kinetics, caba2015activitynet} are converted into free text (either video caption or question-and-answer [QA] formats), thus creating new datasets~\citep{liu2024visual, li2024videochat} and benchmarks~\citep{yu2019activitynet, li2024mvbench}. %In order to train and evaluat MLLMs' action understanding, people change existing action understanding datasets \citep{kay2017kinetics, caba2015activitynet, goyal2017something} to video caption or question \& answer (QA) datasets \citep{liu2024visual, li2024videochat} or benchmarks \citep{yu2019activitynet, li2024mvbench}. 
Those free-text formats offer great flexibility and generalization across datasets, but also introduce limitations in model learning, evaluation, and application perspectives. For the model learning, directly predicting the action name or choosing from some randomly selected candidates (Figure~\ref{fig:ek100-mqa}) prevents the model from learning the full action distributions and contrasting fine-grained actions~\citep{xiao2021next} explicitly. For the model evaluation, free text output makes MLLMs unable to directly compare with previous action task specialized models \citep{ramachandran2025well}. For example, EPIC-KITCHENS-100 \citep{damen2022rescaling}, which is one of the largest and most challenging action datasets, has around four thousand actions. MLLMs may not always predict an action that has an exact match in those action types, as we cannot put all the action types inside MLLMs' context prompt to let MLLMs select. This further limits the applications that require structured actions (e.g., behavior analysis with ethograms~\citep{renner2018ethogram}). 

To address these issues, we take inspiration from the hard example mining literature~\citep{shrivastava2016training, madry2017towards} to improve the learning and evaluation of MLLMs. Specifically, for evaluation, we reformulate EPIC-KITCHENS-100~\citep{damen2022rescaling} into a video multiple-choice question \& answer (MQA) task with the correct ground truth action and four difficult incorrect actions, which we call EPIC-KITCHENS-100-MQA. Incorrect choices are filtered by SOTA action recognition models~\citep{chalk2024tim,zhao2023training} instead of humans or closed-sourced LLMs and MLLMs. This specialized model-based hard example mining reveals substantial drops in performance for existing MLLMs, including GPT-4o (Figure~\ref{fig:ek100-mqa}) and thus offers an efficient and challenging framework for evaluating MLLMs' action recognition abilities. We note that the hard example mining approach automatically picks distractors that pose challenges such as temporal order, which were purposefully curated in other benchmarks~\citep{cai2024temporalbench,li2024mvbench}.

To improve MLLMs' fine-grained action understanding, we proposed an action-related MLLM data transformation regime and curated a training dataset that encompasses various aspects of action understanding, such as hard action recognition, detailed captioning, free-form question answering, temporal detection and prior action association. With the training dataset, we propose LLaVAction models. We introduce an action token designed to improve the model's visual information utilization and a two-stage pipeline to output structured actions and fairly compare with other action recognition models. {\color{revision}{These model designs could be naturally extended to different foundational MLLMs (e.g., \cite{liu2024visual,zhu2025internvl3}).}} LLaVAction obtains SOTA performance on {\color{revision}{four}} action recognition datasets (EPIC-KITCHENS-100~\citep{damen2022rescaling}, EPFL-Smart-Kitchen-30~\citep{bonnetto2025epfl}, MECCANO~\citep{ragusa2021meccano} and Animal Kingdom~\citep{ng2022animal}) 
and shows strong generalization ability in comparison to previous SOTA models. LLaVAction outperforms GPT-4o on EPIC-KITCHENS-100-MQA and achieves consistent improvements on ten video MLLM benchmarks {\color{revision}{that require very different action understanding abilities and are}} either in a caption, open-ended, or multi-choice format.
% Our contributions are as follows:
% \begin{itemize}
% \item We develop specialized model-based hard-example mining and build the challenging EPIC-KITCHENS-100-MQA to evaluate MLLMs' fine-grained action understanding.% as well as effective training strategy for fine-grained action recognition.
% \item LLaVAction achieves SOTA performance on EPIC-KITCHENS-100-MQA (21 points better than GPT-4o), SOTA performance on 3 action recognition datasets and gains consistent improvements on 10 additional video MLLM benchmarks.
% \item We contributed a novel data reformulation for action-related video understanding, which covers multiple perspectives of action understanding. Furthermore, a two-stage pipeline and novel action token are introduced to improve MLLMs on action recognition.

% \item Through our attention-based analysis, we show that our hard-example mining largely improves models attention over visual attentions. Furthermore, our action token becomes an effective action related visual summary for models to understand actions.

% % \item LLaVAction model that achieves strong action understanding ability due to the proposed action-related training targets and model design.
% % \item SOTA performance on action recognition benchmarks.
% \end{itemize}

\begin{figure}[t]
    \centering
    \includegraphics[width=\linewidth]{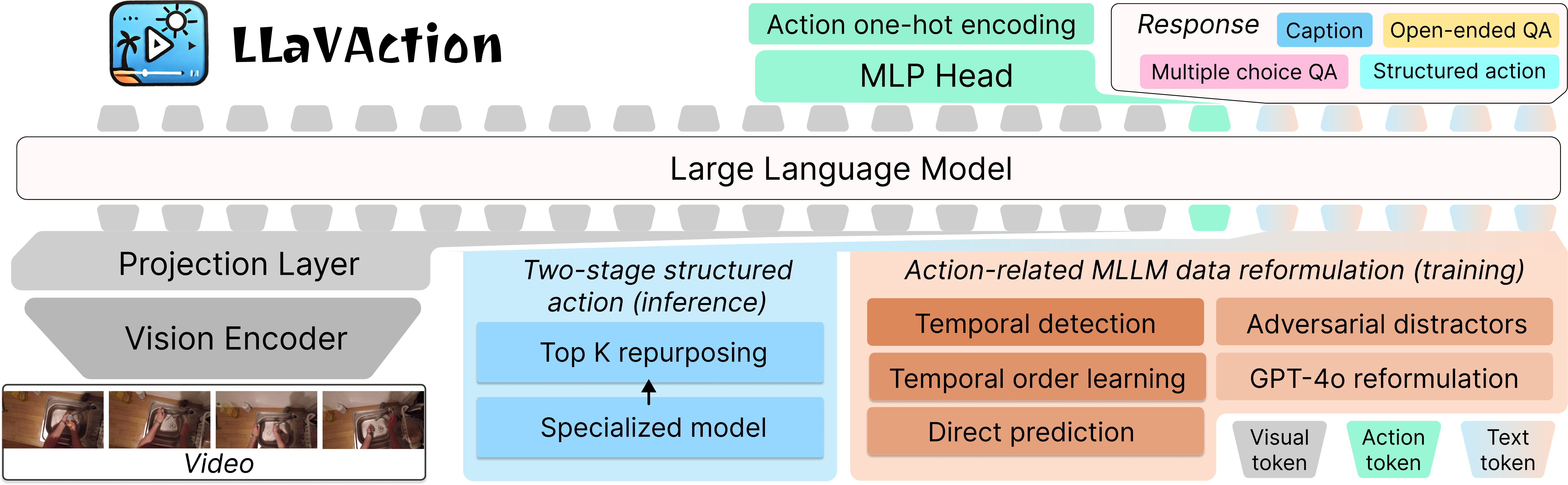}
    \caption{\textbf{LLaVAction pipeline.} Trained with our action-related MLLM reformulated data, LLaVAction outputs captions, action tokens and open-ended and multi-choice QAs. Our two-stage pipeline further enables LLaVAction to output structured action.}
    \vspace{-.3cm}
    \label{fig:llavaction}
\end{figure}

\section{Related Works}

\textbf{Multi-modal large language models.} Multi-modal large language models (MLLMs) are promising generalists~\citep{li2024multimodal}. Early multi-modal models~\citep{tsimpoukelli2021multimodal} mostly performed few tasks or relied on few-shot learning for task generalization.  After the large success of Large Language Models \citep{achiam2023gpt}, multi-modal models appeared that can supplement text with other modalities~\citep{han2024onellm}. Among them, video MLLMs~\citep{li2024llava, zhang2024video} promise robust and scalable solutions to understand and process video data. Our work falls into this direction, aiming at improving MLLMs' action understanding. {\color{revision}{ Action understanding is one of the fundamental abilities in video understanding and has been explored by several recent works with different specific focuses, such as InsTALL~\citep{nguyen2025install} mainly focusing on procedural action planning/prediction, HAIC~\citep{wang2025haic} mainly focusing on detailed action captions and MotionLLM~\citep{chen2024motionllm} mainly focusing on human motion understanding. Instead, our work cares more about fine-grained action contrastiveness.}}

\textbf{MLLM datasets and benchmarks.} Significant efforts have been made to improve training \citep{liu2024visual,li2024videochat} and benchmarking \citep{yue2024mmmu,liu2024mmbench} for MLLMs. Based on the question type, they can be classified as video caption, open-ended question answering and multi-choice question answering (MQA) types. MQA format gains more popularity, especially for benchmarks \citep{xiao2021next,fu2024video}, since it has no need to use another LLM/MLLM to evaluate the model outputs and our EPIC-KITCHENS-100-MQA benchmark falls into this direction. In comparison to the existing MQA benchmarks whose choices are either constructed by humans~\citep{yu2019activitynet,fu2024video} or by closed-source MLLMs~\citep{maaz2023video,ye2024mm}, our benchmark uses action recognition models to efficiently find hard distractors, which is more efficient compared to human generation and is not be limted by closed-source MLLMs' performance. %More importantly, we also extend this specialized model-based hard example mining strategy to construct our fine-grained action understanding training dataset.

\textbf{Action recognition.} Action recognition requires models to predict the action class for a trimmed segment~\citep{shahroudy2016ntu,damen2022rescaling} and is a fundamental task in video understanding~\citep{feichtenhofer2019slowfast,tong2022videomae}. 
% For egocentric vision, action recognition also serves as an important task in many egocentric datasets (e.g., EPIC-KITCHENS-100~\citep{damen2022rescaling}, Ego4D~\citep{grauman2022ego4d}, and HoloAssist~\citep{wang2023holoassist}) due to its importance for applications such as augmented reality and robotics. 
Over the years, many methods have been proposed, yet suffer from fast camera movement, long-term temporal relations, and open vocabulary ability~\citep{damen2022rescaling,grauman2022ego4d}. We focus on MLLMs enhanced with video instruction-tuning~\citep{zhang2024video} to address those challenges.

% \textbf{Vision-centric MLLMs.} The visual performance of MLLMs is likely overestimated due to an over-emphasis of the language-centric tasks in both the model training and the evaluation benchmarks~\citep{tong2024eyes,tong2024cambrian1fullyopenvisioncentric,shangguan2024tomato}.
% There are ongoing efforts to enhance the vision component of MLLMs, such as feeding cropped objects into MLLMs~\citep{shao2024visualcotadvancingmultimodal}, utilizing resampler techniques through visual search~\citep{wu2023vguidedvisualsearch}, and implementing multi-modal Chain of Thought (COT) approaches~\citep{zhang2024multimodalchainofthoughtreasoninglanguage}.

%\textbf{Hard example mining for LLM.} 
%The idea of hard-example mining has been also leveraged in LLM. One example is adversarial filtering. 

\section{Methods}
\label{sec:methods}

We introduce the EPIC-KITCHENS-100-MQA benchmark (Section \ref{sec:ek100mqa}) and a novel MLLM data reformulation paradigm (Section \ref{sec:adversairal_training}), followed by the LLaVAction model designs (Section \ref{sec:our_model}).

\subsection{Hard example mining for MLLM evaluation}
\label{sec:ek100mqa}

Existing MLLMs have shown the ability to understand video content including actions. However, whether MLLMs are good at contrasting fine-grained actions is not clear. Researchers have developed benchmarks to focus on certain aspects of fine-grained actions, such as temporal order \citep{liu2024tempcompass}, which are generated either by human effort or closed-source MLLMs. In this work, we propose to leverage SOTA action recognition models with hard example mining to construct a new benchmark named EPIC-KITCHENS-100-MQA, which is more efficient compared to human generation and is not limited by closed-source MLLMs' performance. More importantly, the proposed hard example mining paradigm can also help MLLMs to enhance fine-grained action understanding (Section~\ref{sec:adversairal_training}) and enable the fair comparisons with specialized models (Section \ref{sec:our_model}).

We use EPIC-KITCHENS-100~\citep{damen2022rescaling} as the data source for our benchmark for the following reasons. Firstly, EPIC-KITCHENS-100 boasts fine-grained action at scale (90K action segments comprising 100h, 4k action types in 100 hours). Secondly, despite numerous models being developed for this dataset, benchmark performance on tasks such as action recognition and segmentation remains far from saturated. Thirdly, the benchmark proves opportunities to compare against specialized models. Importantly, our hard example mining strategy is generalizable and can be applied to any other action understanding dataset. (Section~\ref{sec:action_recognition_results}).

% In terms of the action recognition task, EPIC-KITCHENS-100 contains two types of ground truth annotations for the action segments: one is a compressed version of verb-noun pairs, generated after part-of-speech and clustering; we denote these as 'official keys'. The other ground truth is the narration from the recorders. The narration was not used as the main source of learning in previous methods (e.g., used in pre-training via contrastive learning), as they relied on fixed-vocabulary classifiers~\citep{wu2022memvit,girdhar2022omnivore,zhao2023learning,zhao2023training,chalk2024tim}. However, narrations are the natural learning signals for MLLMs and we argue they are the preferred signals over the official keys as the latter oversimplify language, introducing grammatical and semantic errors (shown in Appendix Figure~\ref{fig:narration_official_key} and Table~\ref{tab:key_narration_comparison}), which might further mislead MLLMs. 

% Therefore, we used the narrations to adapt EPIC-KITCHENS-100's validation set into a video MQA benchmark.
We constructed EPIC-KITCHENS-100-MQA with hard example mining as follows: Let $\mathcal{V} = \{v_1, v_2, ..., v_N\}$ denote the set of video clips. Let $\mathcal{N} = \{n_1, n_2, ..., n_N\}$, $\mathcal{A} = \{a_1, a_2, ..., a_N | a_i \in \mathcal{C}\}$ be their corresponding clip narrations and action labels separately, where $\mathcal{C}$ represents the set of action classes.
For each data sample \( i \), we formulate the MQA task as:
\begin{equation}
    f: (v_i, \mathcal{Q}, \mathcal{O}_i) \mapsto \left[ p_1, p_2, \dots, p_K \right], \text{where} \quad \sum_{k=1}^{K} p_k = 1, \quad p_k \in [0, 1]
\end{equation}
where \( v_i \) is the input sample (e.g., video clip), {\color{revision}{$p_i$ is the probability of picking the i-th option in the MQA as the answer,}} \( \mathcal{Q} \) is the space of possible questions, \( \mathcal{O}_i = \{n_i, \mathcal{D}_i\} \) represents the set of \( K \) answer options, \( n_i \) is the correct narration, and \( \mathcal{D}_i \) represents $K-1$ sampled distractors. These can be sampled randomly from narrations in other action classes:

\begin{equation}
    \mathcal{D}_i^r = \text{Uniform}(\{ n_j \in \mathcal{N} \mid c_j \in \mathcal{C} \setminus \{a_i\} \})
\end{equation}

However, random sampling $\mathcal{D}_i^r$ likely contains trivially wrong answers (Figure \ref{fig:ek100-mqa}). We utilize action recognition models $g: \mathcal{V} \rightarrow (0,1)^{|\mathcal{C}|}$ to find distractors.
% \begin{equation}
%     g: \mathcal{V} \rightarrow (0,1)^{|\mathcal{C}|}
% \end{equation}
For video a specific $v_i$, we obtain the top $K-1$ predicted classes: $\mathcal{C}i = \text{Top}_{K-1}(g(v_i) \setminus \{a_i\})$.
% \begin{equation}
% \mathcal{C}i = \text{Top}_{K-1}(g(v_i) \setminus \{a_i\})
% \end{equation}
The distractor sampling becomes:

\begin{equation}
    \mathcal{D}_i^m = \text{Uniform}(\{ n_j \in \mathcal{N} \mid c_j \in \mathcal{C}_i \})
\end{equation}

The complete set of answers is formed as $\mathcal{O}_i^r = \{n_i\} \cup \mathcal{D}_i^r$ for random sampling or $\mathcal{O}_i^m = \{n_i\} \cup \mathcal{D}_i^m$ for model-based sampling. We used $K = 5$ for our benchmark. {\color{revision}{Moreover, we use the action narrations $\mathcal{N}$ instead of action labels $\mathcal{C}$ to build the choices to avoid implausible texts and confusions (more details are in Appendix \ref{sec:action_label})}} We compared the two sampling strategies. We chose two leading action recognition methods on EPIC-KITCHENS-100, namely, AVION~\citep{zhao2023training} and TIM~\citep{chalk2024tim}. The results indicate that the TIM method consistently produced more challenging distractors for the evaluated MLLMs (Table~\ref{tab:distractors} and qualitative examples Appendix Figure~\ref{fig:choice_comparison}). Consequently, we fixed $g$ with TIM for the EPIC-KITCHENS-100-MQA benchmark. We found that all tested models have a huge performance drop in comparison to the easy setting, which illustrates that MLLMs struggle with fine-grained action recognition when tested with visually or semantically similar actions.

\begin{wraptable}{r}{0.5\linewidth}
\centering
\scriptsize
\vspace{-.35cm}
\begin{NiceTabular}{Wl{45pt} Wc{30pt}Wc{30pt}Wc{30pt}}
\toprule
\shortstack{Choice \\ Selection}    & \shortstack {Random 5 \\ (Easy)} & \shortstack {Avion-Top 5 \\ (Medium)} & \shortstack {TIM-Top 5 \\ (Hard)} \\ \midrule
GPT-4o-mini (07-18)    & 72.0          & 44.2       & 37.4     \\
GPT-4o (08-06)         & 87.6        & 56.7       & 52.2     \\
LLaVA-OV-0.5B  & 59.3           & 37.1          & 32.0       \\
LLaVA-OV-7B    & 68.8           & 33.6          & 28.9     \\
LLaVA-Video-7B & 65.0           & 40.0          & 35.7     \\ 
\bottomrule
\end{NiceTabular}
\vspace{-.15cm}
\caption{\textbf{Comparison between different sources of distractors.} Models were evaluated on either random, AVION or TIM-generated distractors. The values are reported as percent accuracy.}
\vspace{-.4cm}
\label{tab:distractors}
\end{wraptable}

% Note that this test benchmark is distinct from the validation set of EPIC-KITCHENS-100 in two significant aspects. First, it has been reformulated to MQA to facilitate the evaluation of MLLMs. Second, EPIC-KITCHENS-100 employs `official keys' rather than ground truth narrations (as we use). If we were to directly compare with SOTA methods in EPIC-KITCHENS-100, our MLLM can be considered as a refinement-stage that selects the right answer from the predictions of $g$. To make a fair comparison, we also report these results using the official keys in Table~\ref{tab:EPIC-KITCHEN-Original}, even though, as we demonstrate below, the official key is not an ideal text input for MLLMs.

\subsection{Action understanding from multiple perspectives}
\label{sec:adversairal_training}

When training MLLMs on action understanding datasets~\citep{kay2017kinetics}, researchers usually re-formulate those datasets to MLLM-compatible tasks (video caption or question \& answer (QA)), either by directly outputting the original action annotation or reformulating the annotations with closed-source MLLMs. Although this reformulation provides flexibility and generalization across video data even beyond action understanding, fine-grained action differences are not fully explored and hence cause a performance drop on EPIC-KITHCENS-100-MQA. To address this, we expand the previous reformulation regime to encompasses various aspects of action understanding. %In the following, we highlight the methods that we developed that empirically improve performance in action recognition.

\begin{figure}[t]
    \centering
    \includegraphics[width=\linewidth]{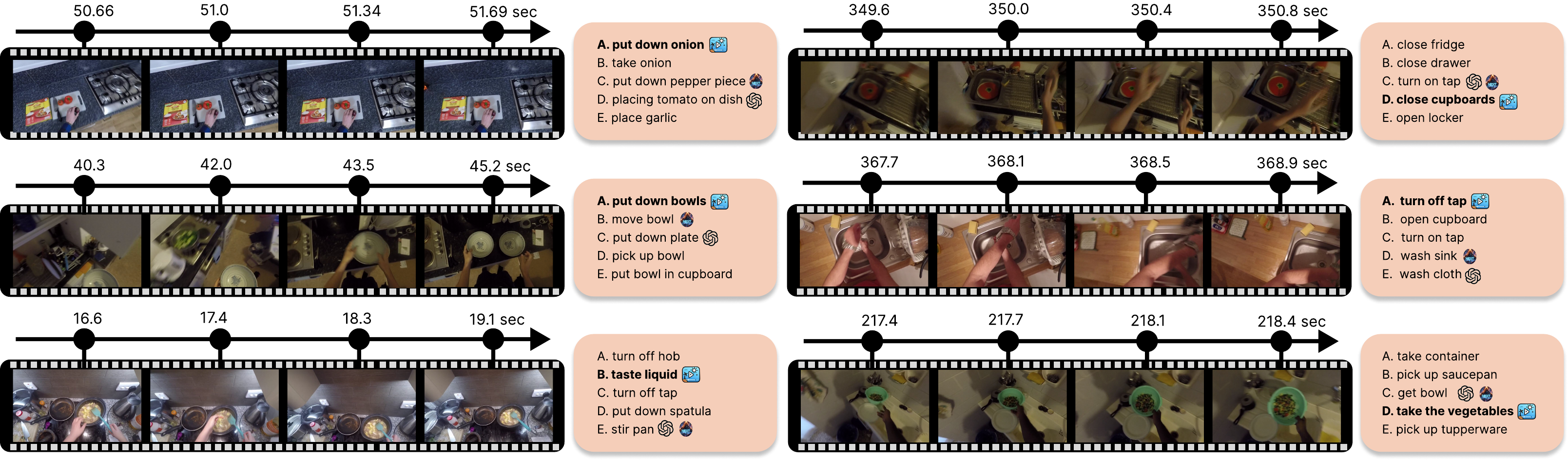}
    \caption{\textbf{Qualitative results.} LLaVAction-7B consistently outperforms GPT-4o and LLaVA-Video-7B when tested on hard distractors. Bold option denotes ground truth, and the icons denote the selection of the models. See also Appendix~\ref{sec:qualitative_examples}.}
    \vspace{-.3cm}
    \label{fig:qualitative}
\end{figure}

\textbf{Adversarial distractors for fine-grained action contrasting.}
Thanks to the effectiveness of our hard example mining paradigm, we can also train MLLMs with `adversarial' distractors that were generated by action recognition models. 
% Unlike for the benchmark, we train MLLMs with both official keys and narrations to serve as data augmentation. To avoid the models from memorizing the position of the ground truth in the options or the order of distractors, we always randomly shuffle the answers. Meanwhile, 
Since the EPIC-KITCHENS-100-MQA benchmark is based on TIM's predictions, training MLLMs to pick the right answers using distractors generated by TIM could lead to over-fitting to TIM's error distributions, resulting in an independent and identically distributed (IID) setting.
{\color{revision}{To avoid methods from obtaining better performance by simply overfitting on TIM's choice combinations}}, we used predictions from AVION to reformulate the EPIC-KITCHENS-100 training set. This provides us with an out-of-distribution (OOD) evaluation on EPIC-KITCHENS-100-MQA (more OOD supports in Appendix~\ref{sec:OOD}). Moreover, we also present the results derived from training with distractors generated by TIM, which is the IID setting and yields the strongest results in EPIC-KITCHENS-100-MQA (Table~\ref{tab:main-results}).

\textbf{Temporal detection for action boundary learning.} Many actions (such as putting something) have clear initiation and conclusion and cannot be explicitly learned by the previous reformulation regime. Therefore, we instruct the model to predict the start and end timestamps of an action sample based on its randomly padded video clip.
Specifically, given a video segment $v_i$ with start timestamp $s_i$ and end timestamp $e_i$, we introduce a fixed temporal padding $\delta = 3$ seconds distributed between start and end. Let $\alpha \sim \text{Uniform}(0, 1)$ be the proportion of padding allocated to the start:
\begin{equation}
    \hat{s}_i = s_i - \alpha\delta, \hat{e}_i = e_i + (1-\alpha)\delta
\end{equation}

We therefore use the new timestamps $\hat{s}_i$ and $\hat{e}_i$ to obtain the padded video segment $\hat{v}_i$. During training, our LLaVAction model takes $\hat{v}_i$ as input and predicts the start and end times as strings (e.g., ``3.20'', ``1.20'') corresponding to the true start and end times of the action in the padded video (prompts are detailed in the Appendix~\ref{sec:temporal_detection}).

\textbf{Temporal order learning with prior actions.} Actions exhibit a certain natural continuity. This temporal aspect of actions can improve predictability. %, particularly when prior actions are known. 
% For instance, in the context of human movement, sequences such as walking, running, or jumping can be anticipated based on the initiation of previous actions. Moreover, 
%In complex tasks such as cooking, the presence of specific steps often indicates the next action to be taken, thereby enhancing predictive modeling. 
% In addition, see Appendix~\ref{sec:prior_action}.
Therefore, we want to leverage prior actions and learn:
%\begin{equation}
$\theta^* = \arg\max_{\theta} \sum_{t=n+1}^{T} \log P_{\theta}(a_t \mid a_{t-1}, \dots, a_{t-n}),$
%\end{equation}
where $\theta^*$ is the optimal set of model parameters we are trying to find, $a_t$ is the current action at time t, $a_{t-1}, \dots, a_{t-n}$ represents the sequence of $n$ previous actions. We set $n = 2$. We modulate this with additional visual instructions to provide prior action information (prompts are detailed in the Appendix~\ref{sec:prior_action}).
% We rely on visual instruction tuning to learn this implicitly. Therefore, we give the prompt, that includes the ground truth actions, such as:\textit{``2.83 seconds ago, you started an action `take paper'. 0.38 seconds ago, you started an action `open bin'. What action are you currently performing? Here are the options of actions you can select''.}
During training, 30\% of the MQAs are provided with prior action information as additional input. 
% During training, 70\% of the time we perform standard MQA and in the remaining 30\%, we perform MQA with previous two actions provided as context.
During evaluation (Table~\ref{tab:additive_ablation}), we can either give the model no prior actions or give the model's own predictions of the previous $n$ actions to formulate as sequential action prediction (SAP).
% We evaluate the effectiveness of this method in two settings: (1) The baseline setting where no prior actions are given. (2) A test time augmentation setting where the models' own predictions of previous $n$ actions are used at test time. Note, we use $n=2$ in both training and testing in the paper and we use ground truth for the previous $n$ actions during training.

% \textbf{Egocentric vs. Allocentric perspective.} MLLM can be assigned with different roles before seeing the video. Since the videos are taken from the first-person perspective, we believed the egocentric perspective aligns better with the LLM pretraining data. Therefore, we switch from the third-person (allocentric) perspective to the first-person (egocentric) perspective to better guide the model. We present the prompts in the Appendix~\ref{sec:perspec_prompt}.

\textbf{Direct prediction.}
Following common practice, we also let the model to directly predict the action descriptions for a given video. The prompt is available in the Appendix~\ref{sec:direct_prediction}.

\textbf{General video understanding with closed-source MLLM-based reformulation.} Although we mainly focus on improving MLLMs' fine-grained action understanding, we do not want to weaken MLLMs' general video understanding. Therefore, we follow previous practice to let the model give a general description of the video or answer video-related questions, where the annotations are obtained from closed-source MLLM GPT-4o. More details are in the Appendix~\ref{sec:gpt_distillation}. As discussed above, we argue that this reformulation alone would not help to gain better fine-grained action understanding. Since GPT-4o itself struggles with our empirically hard distractors, using annotations from GPT-4o alone harms the performance on EPIC-KITCHENS-100-MQA (Table~\ref{tab:distractors}).% benchmark if not trained on MQA.
% Following common practices~\citep{liu2024visual,zhang2024video}, we craft two ways for GPT-4o to annotate. One is to provide captions of the video clips, and the other is to generate open-ended question-answer pairs. 

\subsection{LLaVAction models for better action understanding}
\label{sec:our_model}

With the proposed new MLLM data reformulation paradigm, we propose LLaVAction, which is better at fine-grained action understanding. LLaVAction is developed based on LLaVA series~\citep{li2024llava,zhang2024video}. We further design a learnable action token to enhance the visual information utilization and a two-stage pipeline to output structured action so that we can compare LLaVAction fairly with other SOTA action recognition models (Figure~\ref{fig:llavaction}).

% Thus, we benchmark these as baselines. We present results for both the 0.5 billion (0.5B) and 7 billion (7B) parameter models of LLaVA-OneVision, and the LLaVA-Video-7B, as it does not provide a 0.5B alternative. In all experiments, we started from their released checkpoints~\citep{llava_onevision_0.5B, llava_onevision_7B, llava_video_7B} (Appendix~\ref{sec:licensing}). In the following, we highlight the methods that we developed that empirically improve performance in action recognition.

% \textbf{Adversarial distractors from action recognition models.}
% We ran inference for AVION and TIM on EPIC-KITCHENS-100 using their officially released code and weights to obtain their predictions on both training and validation sets. 
% We sort the predictions by model confidence and keep the top 5 predictions as distractors. To ensure that ground truth is always in the $K$ options in the training, we remove the least confident prediction in the set and replace it with the ground truth (either with ground truth narration or official key, as appropriate for the experiment). To avoid the models from memorizing the position of the ground truth in the options or the order of distractors, we always shuffle the answers.

\textbf{Enhancing visual information utilization with action token.} 
Most MLLMs rely only on the language prediction of the next token to train the model and extract information from visual tokens~\citep{liu2024visual}. However, recent findings suggest that this training strategy decreases the importance of vision tokens in late layers of MLLMs~\citep{zhang2025llavaminiefficientimagevideo, liu2024multistagevisiontokendropping}. Therefore, we designed an intermediate supervision of visual tokens. Specifically, we added a learnable action token into the input tokens, {\color{revision}{which is analogically similar to the CLS token in the VIT model~\citep{dosovitskiy2020image} that has been shown the ability in grasping image content~\citep{wang2025videollamb}}} The order of the input tokens was system text tokens, visual tokens, learnable action token and instruction text tokens; through causal attentions of the LLM backbone this enables the action token to integrate action information from visual tokens and then contribute to the subsequent language tasks. 
Let's denote the hidden states at the final layer of the MLLM as:
\begin{equation}
    \left\langle H^q_1, \cdots, H^q_k, H^v_1, \cdots, H^v_{l_v}, h^a, H^q_{k+1}, \cdots, H^q_{l_q}\right\rangle
\end{equation}
$H^v_i \in \mathbb{R}^d$ are the hidden states corresponding to visual tokens, $h^a \in \mathbb{R}^d$ denotes the hidden state of the learnable action token, $H^q_i  \in \mathbb{R}^d$ denotes the text tokens,  $d$ denotes the hidden dimension of the LLM. $l_q$ and $l_v$ denote the length of text tokens and length of visual tokens respectively.
We apply three classification heads on top of the hidden state \( h^a \) to predict nouns, verbs, and actions separately and use cross-entropy loss to train the classifiers, in the belief that the classification training could guide the action token to learn better extract action-related visual information. {\color{revision}{Note that our action token only serves as an additional learning objective and is not used to give the final prediction. During inference, or when training with tasks that have no clear action labels (e.g., video captioning), we can simply compute the text generation loss.}}

\begin{wraptable}{r}{0.4\linewidth}
\centering
\scriptsize
\vspace{-.4cm}
\begin{tabular}{lll}
\toprule
Methods        & 8 f & 16 f \\
\midrule
zero-shot GPT-4o         & 52.2     & N/A       \\
zero-shot GPT-4o-mini    & 37.4     & N/A       \\
zero-shot LLaVA-Video-7B & 35.7     & 34.8      \\
zero-shot LLaVA-OV-7B    & 28.9     & 30.5         \\
zero-shot LLaVA-OV-0.5B  & 32.0       & 31.6        \\ \midrule
LLaVAction: LLaVA-Video-7B & 71.7        & \textbf{73.4} \\
LLaVAction: LLaVA-OV-7B    & 71.3        & 72.3        \\
LLaVAction: LLaVA-OV-0.5B  & 64.8        & 65.4  \\
\bottomrule
\end{tabular}
\vspace{-.2cm}
\caption{\textbf{Comparison on EPIC-KITCHENS-100-MQA.} Columns represent the number of frames used for testing. Percent accuracy is shown.}
\vspace{-.3cm}
\label{tab:main-results}
\end{wraptable}

\textbf{Two-stage structured action prediction pipeline.} 
MLLMs directly output free texts, which may be hard to find an exact match with the action labels in the dataset. One could put all the possible actions in the question prompt and let MLLMs choose, but this will become infeasible when the number of action types increases (e.g., around 4k action types in EPIC-KITCHENS-100). This limitation prevents MLLMs from fairly comparing with action recognition models and constrains MLLM from applying to applications that require structured actions. To this end, LLaVAction designs a two-stage pipeline when it needs to output structured actions. Similar to hard example mining (Section \ref{sec:ek100mqa}), we use action recognition models' predictions to filter out easy actions. The difference is that we directly take the top K confident actions without adding the Ground Truth (GT) action. The two-stage pipeline will lower the upper-bound performance MLLM can obtain since the GT action may not be in the top K prediction. However, the K value can control the trade-off between the upper-bound performance and the number of actions the MLLM needs to contrast. {\color{revision}{Note that our two-stage pipeline is only needed for datasets and applications that require structured actions, and is only applied at the inference stage. For the open-vocabulary tasks, LLaVAction can directly take the video and questions and give an open-ended answer. For training and benchmark construction, we always include the GT action into the choices to avoid misalignment.}}

\section{Experiments}
\vspace{-3pt}
%\subsection{Evaluation benchmarks and details}

We conducted evaluations for the LLaVAction models on a wide range of benchmarks to show the models' strong action understanding ability. We used LMMs-Eval~\citep{zhang2024lmms} to evaluate LLaVAction on different MLLM benchmarks. We used eight frames for distractor experiments (Table~\ref{tab:distractors}), 8 or 16 frames for our main results on EPIC-KITCHENS-100-MQA (Table~\ref{tab:main-results}) and 16 frames for ablation studies (Table~\ref{tab:leave_one_ablation}), so that we can limit the cost of calling closed-sourced models during comparison. Meanwhile, we use 32 frames for EPIC-KITCHENS-100 (Table~\ref{tab:EPIC-KITCHEN-Original}) and 64 frames for the model reported in MLLM benchmarks (Table~\ref{tab:additional-video-benchmarks}).

\subsection{Implementation and training details}

We trained LLaVAction with three open-source baseline variants, LLaVA-Video-7B, LLaVA-OV-7B and LLaVA-OV-0.5B. 
% then built a series of methods that empirically boosted performance (see Section \ref{sec:our_model}). In our EPIC-KITCHENS-100-MQA benchmark, ground truth narration is within the 5 options. In addition to ground truth, the distractors are generated from TIM with its highest confidence predictions (see Section~\ref{sec:ek100mqa}).
% For training, we experiment with different sources of distractors for MQA (see Section \ref{sec:adversairal_training}). We study the validation accuracy when we (1) use random sampling for distractors, (2) use AVION generated distractors (highest confidence predictions), and (3) use TIM generated distractors. Different from the EPIC-KITCHENS-100-MQA benchmark, we also include the official key choices as an additional data augmentation for the model to be aware of different representation of actions instead of only using the narration choices.
% In addition to video MQA (Section~\ref{sec:our_model}) we added four auxiliary methods in our training pipeline: (1) direct prediction, (2) temporal detection, (3) GPT-4o distilled video caption~(Appendix Figure~\ref{fig:llavaction_caption_task}), and (4) GPT-4o distilled video open-ended question answering. 
With the proposed new MLLM data reformulation paradigm, we re-annotated EPIC-KITCHENS-100's training data, in total contributing to 530K annotated video-language pairs to train the LLaVAction models.
% Note that our techniques, such as visual token supervision, learning from prior actions, and changing prompt perspectives, do not require us to augment with additional training data.
The 7B and 0.5B models were trained for 12 and 11 hours on 32 GH200 GPUs, respectively.
Across all experiments and all baseline models, we set gradient accumulation to 2, batch size to 64 and total epochs to 2. Following LLaVA-Video, the MLP connector, LLM backbone (Qwen2-0.5B,7B~\citep{wang2024qwen2}) and the visual encoder (SigLIP-384~\citep{zhai2023sigmoidlosslanguageimage}) were trained. The learning rate was $2\text{e-}6$ for the vision encoder and $1\text{e-}5$ for the rest (see~\citet{zhang2024video} for additional details). Training on EPIC-KITCHENS-100 data alone might result in over-fitting. Therefore, we use data replay~\citep{li2024llava} to aid in generalization of the model. Thus, we mix with the training data of LLaVA-Video, namely LLaVA-Video-178K.

% \section{Results}
% \label{sec:results}

% We focus on evaluating and improving leading MLLMs' abilities for recognizing human actions in complex scenes. We focus on fine-grained, multi-tasking actions that occur in kitchen activities. Unlike existing MLLMs that leverage annotations from either distillation from powerful MLLMs or human annotations, we also leverage hard examples selected by leading action recognition models as well as proposing novel methods to improve MLLMs on this task. 

% First, we developed the EPIC-KITCHENS-100-MQA benchmark (Section~\ref{sec:methods}).
% Specifically, we used TIM to generate answers (Table~\ref{tab:distractors}) and used this as the OOD test-set when evaluating our methods (Tables~\ref{tab:main-results} and ~\ref{tab:additive_ablation}). We also trained a model with TIM distractors, thus representing IID (top-line) performance (Table~\ref{tab:additive_ablation}). 

\begin{wraptable}{r}{0.4\linewidth}
\centering
\scriptsize
\vspace{-.35cm}
\begin{tabular}{lll}
\toprule
         LLaVA-Video-7B  $\Rightarrow$ LLaVAction-7B  \\ 
\midrule
 \textbf{OOD Setting}:\\
Zero-shot   & 34.8              \\
+ GPT-4o-based reformulation   & 21.9              \\ 
+ Random distractors  & 55.0              \\ 
\midrule
Adversarial distractors (AVION)   & 64.4             \\ 
+ Temporal Detection  & 65.2              \\

+ Action token  &  69.1             \\

+ GPT-4o-based reformulation    &  71.5             \\
+ Direct Prediction     &  73.6   \\
+ Temporal order learning    &  73.4      \\
+ Temporal order learning w/ SAP              &  \textbf{74.1}  \\
\midrule
\textbf{IID Setting}:\\
+ Adversarial distractors (TIM)     & 76.3 \\
+ Adversarial distractors (TIM) w/ SAP       &  77.0 
\\
\bottomrule        
\end{tabular}
\vspace{-.2cm}
\caption{\textbf{LLaVAction additive ablations on EPIC-KITCHENS-100-MQA}. Techniques are gradually added to achieve the final model. SAP denotes sequential action prediction during inference. Percent accuracy is shown.}
\vspace{-.25cm}
\label{tab:additive_ablation}
\end{wraptable}

\subsection{Results on EPIC-KITCHENS-100-MQA}

EPIC-KITCHENS-100-MQA contains hard distracting choices and is excellent to evaluate MLLMs' fine grained action understanding. We report results for the MLLMs comparison (Table~\ref{tab:main-results}), LLaVAction additive ablations (Table~\ref{tab:additive_ablation}), and leave-one-out ablations (Appendix~\ref{sec:leave_one_out}) on EPIC-KITCHENS-100-MQA.
We start with MLLM comparisons (Table~\ref{tab:main-results}). 
%We evaluated closed-source GPT-4o and GPT-4o-mini, and open-source models LLaVA-Video-7B, LLaVA-OV-7B, and LLaVA-OV-0.5B. 
%GPT-4o from OpenAI performs best under the zero-shot setting (52.2). Of the open-source models, LLaVA-Video-7B performs the best (35.7).
LLaVAction models perform much better than baselines and also obtain a 21-point improvement over GPT-4o (running GPT-4o beyond 8 frames is cost-prohibitive). Next, we verified LLaVAction's improvements do not just come from adding in-domain data by additive ablations (Table~\ref{tab:additive_ablation}). Since both our MLLM data reformulation paradigm (Section \ref{sec:adversairal_training}) and LLaVAction model design (Section \ref{sec:our_model}) contribute to the final performance, we ablate them together. We can see training with adversarial distractors (AVION) results in the largest improvement (9.4 points). Addition of the action token gives the second most improvements (3.9 points improvement).
Meanwhile, 
% \textbf{Training with adversarial distractors (AVION).}
% We compare LLaVA-Video-7B training with random distractors and adversarial distractors using AVION's predictions. There is a 9.4 difference in accuracy, showing the effectiveness of training against adversarial distractors.
% \textbf{Perspective Prompt (Egocentric).}
% When we fix the distractors from random sampling, using the egocentric prompt gives a 0.5 point improvement over using the allocentric prompt.
% \textbf{Effectiveness of action token.}
% Inspired by recent findings suggesting that the importance of vision tokens decreases in MLLM in late layers \citep{zhang2025llavaminiefficientimagevideo}, we explored ways to improve vision tokens. Instead of removing redundant vision tokens, we aimed to add an action token and intermediate supervision, which is detailed in Section~\ref{sec:methods}, which gives an additional 3.9 points improvement.
% \textbf{Impact of closed-source MLLM-based reformulation.}
% Closed-source MLLM-based reformulation is widely used in previous MLLM training. However, 
we note that simply fine-tuning LLaVA-Video-7B with the previous MLLM data reformulation paradigm (GPT-4o-based reformulation) results in a performance degradation (35.7 to 21.9). Based on the fact that it gives a meager 2.4 performance boost when we combine it with MQA task using AVION distractors, we believe this is a sign of catastrophic forgetting of MQA capability. Furthermore, directly predicting the action (direct prediction) and using contextual prior actions (temporal order learning w/ SAP) result in 2.1 and 0.5 points improvements, respectively.
% \textbf{Effectiveness of direct prediction.}
% With direct prediction, the model learns to directly predict the action without picking from the given options and video frames. This gives an additional 2.1 points improvement.
% \textbf{Temporal order learning with prior actions.}
% Actions have the sequential nature. We expect that the model can use contextual information such as the prior two actions as a hint for better accuracy. Table~\ref{tab:additive_ablation} shows that sequential action prediction can obtain a 0.5 point improvement.
In summary, the combination of our MLLM data reformulation paradigm and model design greatly improves the performance of the base LLaVA-Video-7B model from 34.8 to 74.1 accuracy in the OOD setting and to 77.0 in the IID setting. %If we consider IID setting and use TIM adversarial distractors during training, the model performance can further reach to 77.0.
% \vspace{-3pt}
\subsection{Results on action recognition benchmarks}
\label{sec:action_recognition_results}
% \vspace{-1pt}

With our two-stage structured action prediction pipeline (Section \ref{sec:our_model}), LLaVAction can be fairly compared with other action recognition models (Table~\ref{tab:EPIC-KITCHEN-Original}). To assess the generalization of LLaVAction, {\color{revision}{we tested on three datasets, which were carefully selected to exclude data used in pretraining while covering different domains.}}

\begin{wraptable}{r}{0.4\linewidth}
\centering
\scriptsize
\vspace{-.3cm}
\begin{tabular}{ll}
\toprule
Methods & Acc. \\ \midrule
{\color{revision}{IPL~\citep{wang2021interactive}}} & {\color{revision}{41.0}} \\
LaViLa~\citep{zhao2022learningvideorepresentationslarge} & 51.0 \\
TAdaFormer-L/14~\citep{huang2023temporallyadaptivemodelsefficientvideo} & 51.8 \\
LVMAE~\citep{gundavarapu2024extendingvideomaskedautoencoders} & 52.1 \\
M\&M~\citep{xiong2022mmmixmultimodalmultiview}                & 53.6 \\
AVION~\citep{zhao2023training}                               & 54.4 \\ 
TIM~\citep{chalk2024tim}                                  & 56.4 \\ \midrule
%Ours, LLaVA-OV-7B w/ official key  & - \\
%Ours, LLaVA-OV-7B w/ GT narration  & 61.8 \\
%\hline
%Ours, LLaVA-OV-0.5B w/ official key  & - \\
%Ours, LLaVA-OV-0.5B w/ GT narration  & 56.3 \\
%\hline
% Ours, LLaVAction-7B (16f) w/ official key            & 57.8 \\ 
% Ours, LLaVAction-7B (16f) w/ GT narration  & 63.1 \\ 
Ours, LLaVAction-7B w/ action label            & \textbf{58.3} \\ 
Ours, LLaVAction-7B w/ action narration  & \textbf{63.2} \\ 
\bottomrule
\end{tabular}
\vspace{-.15cm}
\caption{\textbf{Action recognition on EPIC-KITCHENS-100.} Top-1 accuracy on action classification. For specific verb-noun performance see Figure~\ref{fig:llavaction_analysis}.}
\vspace{-.25cm}
\label{tab:EPIC-KITCHEN-Original}
\end{wraptable}

\textbf{LLaVAction achieves SOTA on EPIC-KITCHENS-100.}
Following common practice, we report the performance on EPIC-KITCHENS-100's validation set. We report performance in two settings ('w/ action label' and 'w/ action narration'), which differ in the candidate choice generation. For the `w/ action label' setting, we directly concatenated verb and noun action classes to obtain choices, which could produce implausible text. For example, the noun class for coffee maker represents `maker:coffee' in the noun definition. Meanwhile, `pour into' is simplified as `pour' in the verb definition, which could generate implausible text such as `pour pot' that should be `pour into pot'. Based on those observations, we also report the 'w/ action narration' setting where we used the action narration of the corresponding video clip (more details in Appendix \ref{sec:action_label}). We empirically observe that we get better results scaling top-K from $5$ to $20$. Therefore, we train and evaluate LLaVAction with TIM's top 20 action predictions. LLaVAction achieves SOTA on EPIC-KITCHENS-100 under both settings.

\begin{wraptable}{r}{0.6\linewidth}
\centering
\scriptsize
\vspace{-.3cm}
\setlength{\tabcolsep}{3pt}
\begin{tabular}{lccc}
\toprule
Methods & Acc. & Head Acc. & Tail Acc. \\ 
\midrule
\multicolumn{4}{c}{\textit{Zero-shot generalization {\color{revision}{of the second-stage model}}}} \\
\midrule
AVION~\citep{zhao2023training}                 & 19.3   & 21.2   & 8.6 \\ 
LLaVA-Video-7B~\citep{zhang2024video}          & 22.5   & 22.9   & 18.8 \\
Ours, LLaVAction-7B                           & \textbf{36.2}   & \textbf{38.1}   & \textbf{24.6} \\
\midrule
\multicolumn{4}{c}{\textit{Trained model}} \\
\midrule
VideoMAE~\citep{tong2022videomae}              & 37.5   & 41.1   & 16.6 \\
Multi-modal VideoMAE~\citep{bonnetto2025epfl}  & 40.0   & 43.6   & 19.4 \\
Ours, LLaVAction-7B                           & \textbf{46.6}   & \textbf{49.7}   & \textbf{27.0} \\
\bottomrule
\end{tabular}
\caption{\textbf{Action recognition on EPFL-Smart-Kitchen-30.} LLaVAction outperforms prior methods in the zero-shot and finetuned setting.}
\vspace{-.2cm}
\label{tab:EPFL-Smart-Kitchen-30}
\end{wraptable}

%\textbf{Comparison on EPFL-Smart-Kitchen-30.}
\textbf{LLaVAction generalizes well to other datasets.} We tested LLaVAction on two recent action recognition benchmarks -- one testing generalization for a different cooking dataset (EPFL-Smart-Kitchen-30,~\citep{bonnetto2025epfl}) and for a different domain (tool assembly, Meccano,~\citep{ragusa2021meccano}). EPFL-Smart-Kitchen-30 has 30 verbs and 46 nouns in common with EPIC-KITCHENS-100, which enables us also to compare with specialized models (such as AVION) for zero-shot generalization {\color{revision}{of second-stage models}}. We used the multi-modal VideoMAE's top 5 predictions in EPFL-Smart-Kitchen-30 to generate the MQAs for LLaVAction-7B and LLaVA-Video-7B. To fairly compare with the specialized model AVION, we also use those top 5 predictions to filter AVION's predicted action logits. LLaVAction obtained better zero-shot accuracy than AVION and LLaVA-Video (Table~\ref{tab:EPFL-Smart-Kitchen-30}). Most importantly, the zero-shot LLaVAction even obtained similar overall performance to the trained VideoMAE model and better tail action accuracy (Tail Acc.). When finetuning LLaVAction obtained SOTA performance (Table~\ref{tab:EPFL-Smart-Kitchen-30}).

On Meccano we used SlowFast~\citep{feichtenhofer2019slowfast} to generate hard distractors and finetuned LLaVAction. Even when trained for only one epoch, LLaVAction obtained 51.7 top-1 accuracy, beating SlowFast’s 42.8.

\begin{wraptable}{r}{0.42\linewidth}
\centering
\scriptsize
\vspace{-.4cm}
{\color{revision}{
\begin{tabular}{lc}
\toprule
Methods & Jaccard Acc. \\ \midrule
VideoMAE~\citep{mamooler2025fine} & 53.1 \\ \midrule
% \multicolumn{2}{c}{\textit{LLaVA-Video-7B}} \\ \midrule
LLaVA-Video-7B (Zero-shot) & 30.5 \\
LLaVA-Video-7B (Random choice) & 46.7 \\
Ours, LLaVA-Video-7B                & 61.0 \\ \midrule
% \multicolumn{2}{c}{\textit{InternVL3-8B}} \\ \midrule
InternVL3-8B(Zero-shot) & 28.1 \\
InternVL3-8B(Random choice) & 43.9 \\
Ours. InternVL3-8B        & 58.7 \\

\bottomrule
\end{tabular}
\vspace{-.2cm}
\caption{\textbf{Action recognition on Animal Kingdom.}}
\label{tab:animalkingdom}
}}
\vspace{-.2cm}
\end{wraptable}

{\color{revision}{

Our training pipeline and model design is not limited to any particular domains or base MLLMs. To further support that, we tested our method on a very different domain, animal fine-grained behavior understanding. Specifically, we tested on the Animal Kingdom dataset~\citep{ng2022animal}, which has 140 fine-grained actions and at most 12 actions can happen at the same time (i.e., multi-classification task). We adapted both LLaVA-Video-7B and InternVL3-8B~\citep{zhu2025internvl3} and trained with hard examples generated from VideoMAE~\citep{mamooler2025fine} to serve as our methods; for two base models (LLaVA-Video-7B and InternVL3-8B) we compared to zero-shot and training with randomly generated options. Our methods, either using LLaVA-Video-7B or InternVL3-8B as the base MLLMs obtained much better performance (61.0/58.7) compared to the original models (30.5/28.1) and models finetuned with random choices (46.7/43.9) (Table \ref{tab:animalkingdom}). Furthermore, our method beats the baseline (VideoMAE) for both base models.

}}

\subsection{Results on other MLLM Benchmarks}

Apart from comparing with action recognition models, we want LLaVAction to keep general video understanding abilities and also improve fine-grained action understanding on other zero-shot MLLM benchmarks. Therefore, we tested LLaVAction-7B on 13 MLLM benchmarks that test various MLLM video understanding abilities. The evaluated benchmarks consist of two video caption benchmarks, five open-ended Q\&A benchmarks and six multi-choice Q\&A benchmarks. LLaVAction-7B outperforms LLaVA-Video-7B on 10 benchmarks, indicating the enhanced video understanding ability of our model (Table~\ref{tab:additional-video-benchmarks}).

\begin{table}[t]
\centering
\scriptsize
\setlength{\tabcolsep}{3pt}
\begin{tabular}{lcc|ccccc|cccccc}
% \begin{NiceTabular}{Wl{90pt} Wc{12pt}Wc{12pt}|
% Wc{12pt}Wc{12pt}Wc{12pt}Wc{12pt}Wc{0pt}|
% Wc{12pt}Wc{12pt}Wc{12pt}Wc{12pt}Wc{12pt}Wc{12pt}}
\toprule
  & \multicolumn{2}{c}{\textit{Caption}} & \multicolumn{5}{c}{\textit{Open-ended Q\&A}} & \multicolumn{6}{c}{\textit{Multi-choice Q\&A}} \\
\midrule
  &  \rotatebox{90}{\shortstack[l]{VDC \\ \citep{chai2024auroracap}}} & \rotatebox{90}{\shortstack[l]{VideoDC\\ \citep{chen2024sharegpt4video}}}
  &  \rotatebox{90}{\shortstack[l]{VideoEval-Pro\\ \citep{ma2025videoeval}}} & \rotatebox{90}{\shortstack[l]{ActNet-QA\\ \citep{yu2019activitynet}}} & \rotatebox{90}{\shortstack[l]{VideoChatGPT\\ \citep{maaz2023video}}} & \rotatebox{90}{\shortstack[l]{CVRR\\ \citep{khattak2025good}}}  & \rotatebox{90}{\shortstack[l]{TempCompass\\ \citep{liu2024tempcompass}}}
  &  \rotatebox{90}{\shortstack[l]{EgoSchema\\ \citep{mangalam2023egoschema}}} &  \rotatebox{90}{\shortstack[l]{MVBench\\ \citep{li2024mvbench}}}  & {\rotatebox{90}{\shortstack[l]{VideoMME \\ wo/w-subs\\ \citep{fu2025video}}}} &   \rotatebox{90}{\shortstack[l]{LongVideoBench\\ \citep{wu2024longvideobench}}} & \rotatebox{90}{\shortstack[l]{NextQA\\ \citep{xiao2021next}}}   & \rotatebox{90}{\shortstack[l]{PerceptionTest\\ \citep{patraucean2023perception}}}   \\  
\midrule
\textbf{Closed-source models} & \\
\midrule
GPT-4V \citep{achiam2023gpt}            & -  & -  & - & - & -  & 70.8 & -   & -  & 43.5 & 59.9/63.3 & 61.3  & - & -      \\  
GPT-4o \citep{openai2024gpt4ocard}      & -  & -  & 34.2 & - & -  & - & -   & -  & - & 71.9/77.2 & 66.7  & - & -      \\  
Gemini-1.5-Flash \citep{team2023gemini} & -  & -  & 35.1 & - & -  & - & -   & 65.7 & - & 70.3/75.0 & 61.6  & - & -      \\  
Gemini-1.5-Pro \citep{team2023gemini}   & 41.7  & -  & 39.3 & - & -  & - & -   & 72.2 & - & 75.0/81.3 & 64.0  & - & -      \\  
\midrule
\textbf{Open-source models} & \\
\midrule
LongVA-7B \citep{zhang2024long}         & 34.5  & -  & 16.5 & 50.0 & -  & - & -   & - & - & 52.6/54.3 & -  & 68.3 & -      \\ 
mPLUG-Owl3 \citep{ye2024mplug}          & 38.9  & -  & -    & -    & -  & - &  34.4   & - & 54.5 & 59.3/68.1 & 52.1 & 78.6 & - \\
VideoChat2-7B \citep{li2024mvbench}     & 36.5  & -  & -    & 49.1 & -  & 25.8 & 38.5   & - & \underline{60.4} & 42.3/54.6 & -  & 78.6 & -      \\ 
VideoLLaMA2-7B \citep{cheng2024videollama}& - & -  & -      & 53.0 & -  & 21.6 & 32.2  & 51.7 & 54.6 & 47.9/50.3 & -  & - & 51.4      \\ 
LLaVA-OV-7B \citep{li2024llava}           & 38.8 & -  & -   & 56.6 & -  & - & -  & \textbf{60.1} & 56.7 & 58.2/61.5 & 56.5 & 79.4 & 57.1 \\

LLaVA-Video-7B \citep{zhang2024video}     & \underline{39.0} & \textbf{3.44}  & \underline{25.7} & \underline{66.0} & \textbf{3.04}  & \underline{51.3} & \underline{66.0}  & 57.3 & 58.6             & \underline{63.3}/\underline{69.7} & \underline{58.2}       & \textbf{83.2}    & \underline{67.9} \\ 
\textbf{Ours, LLaVAction-7B}           & \textbf{40.2} & \underline{3.34}  & \textbf{26.1} & \textbf{66.9} & \underline{3.01}  & \textbf{55.6} & \textbf{66.1}  & \underline{59.0} & \textbf{61.1} & \textbf{63.9}/\textbf{71.4}    & \textbf{58.6}    & \underline{82.8} & \textbf{70.2}       \\ 
\midrule
Relative improvement of ours\\ over the baseline LLaVA-Video-7B         & {\color{ForestGreen} +1.2} & {\color{olive} -0.1}  & {\color{ForestGreen} +0.4} & {\color{ForestGreen} +0.9} & {\color{olive} -0.03}  & {\color{ForestGreen} +4.3} & {\color{ForestGreen} +0.1}             & {\color{ForestGreen} +1.7} & {\color{ForestGreen} +2.5} & {\color{ForestGreen} +0.6}/{\color{ForestGreen} +1.7} & {\color{ForestGreen} +0.4}  & {\color{olive} -0.4} & {\color{ForestGreen} +2.3} \\
\bottomrule
\end{tabular}
\vspace{-.1cm}
\caption{\textbf{Performance on other MLLM benchmarks} that contain human actions. Please note, we are not claiming SOTA, we are noting that we can improve performance over our baseline open-source model (LLaVA-Video-7B \citep{zhang2024video}). We also show sub-task performances in Appendix~\ref{sec:additional_benchmarks_2}. We show top-performance closed-source models for reference. Top open-source models are shown in bold, and the second-best are underlined.}
\vspace{-.3cm}
\label{tab:additional-video-benchmarks}
\end{table}

\subsection{Attention-based analysis}
\label{attention_analysis}

% Hard example mining with specialized action models enables us to better train and evaluate MLLMs' fine-grained action understanding. Meanwhile, action token design helps LLaVAction to better utilize action-related information. Those two innovations have shown the most quantitative performance boosts on EPIC-KITCHENS-100-MQA (Table \ref{tab:additive_ablation}). In this section, we aim to analyze their impacts in an explainable way, followed by a summary of the whole work.

\begin{figure}[t]
    \centering
    \includegraphics[width=.9\linewidth]{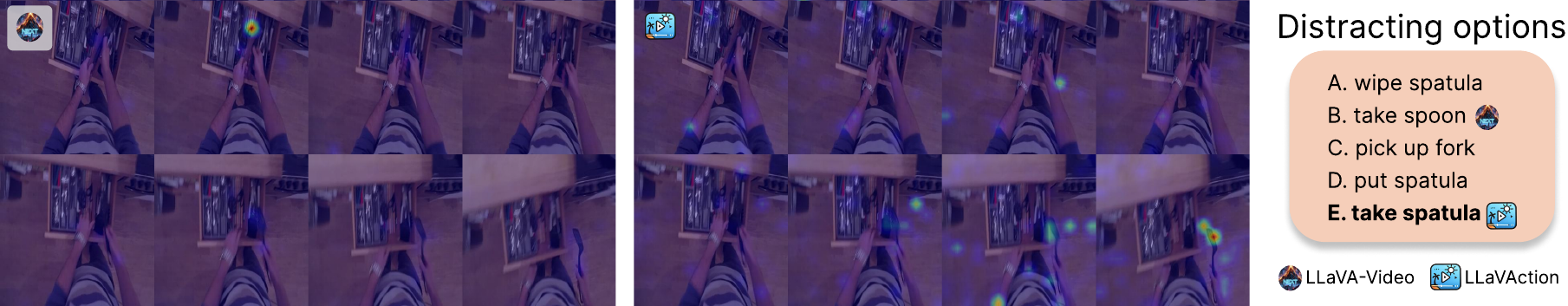}
    \vspace{-.1cm}
    \caption{\textbf{Qualitative attention for one clip.} Anecdotally, LLaVA-Video mainly attends to the wooden spatula that is placed in the drawer, LLaVAction also attends to the arms and, correctly, the plastic spatula that is being taken. We quantify visual-text token correlations in the main text. }
    \vspace{-.1cm}
    \label{fig:attention}
\end{figure}

We sought to analyze the impact of action-related training and LLaVAction model design in an interpretable manner. Following the approach in \citep{zhang2025flexselect}, we employed token attention analysis to understand model behavior (Figure \ref{fig:attention}). We computed average text-visual token correlations for both LLaVA-Video-7B and LLaVAction-7B using the EPIC-KITCHENS-100-MQA dataset. For fair comparison, we fine-tuned LLaVA-Video-7B with randomly generated answer choices. Analysis Methodology: We first computed the text-visual attention tensors of size $N\times T \times V$, where $N$ is the number of data samples, $T$ is the number of text tokens and $V$ is the number of visual tokens. We then calculated the maximum across the text token dimension, followed by computing mean and 90th percentile values to estimate text-visual correlations.

Text-Visual Correlation Results: LLaVA-Video-7B achieved mean and 90th percentile values 0.00476 and 0.0104, respectively, while LLaVAction-7B achieved mean and 90th percentile values 0.00769 and 0.0175, respectively. This indicates that LLaVAction attends more strongly to visual cues compared to LLaVA-Video, likely due to our hard example mining strategy.

Action Token Analysis: We computed the text-action token attention tensors of size $N\times T \times 1$, calculating the maximum over the text dimension followed by the mean to estimate text-action correlations. LLaVAction demonstrates an average text-action token correlation of 0.143, significantly higher than its text-visual correlation (0.00769). Notably, 99\% of visual tokens exhibit lower text correlations compared to action tokens. Since action tokens exclusively attend to visual tokens, this suggests they effectively aggregate visual information relevant to fine-grained actions, enhancing LLaVAction's action understanding capabilities.

%They computed the text-visual token correlations using attention and identified that the 19th layer of the LLaVA model has the best "Needle-in-Haystack" performance. Therefore, we first computed the text-visual token correlations of 19th layer for both LLaVA-Video-7B and LLaVAction-7B to examine the influence of hard distractor training (Section \ref{sec:distractor_analysis}). To make a fair comparison, we finetune LLaVA-Video-7B with randomly generated choices. Next, we computed the text-action token correlations analogically to explore the effect of the designed action token (Section \ref{sec:token_analysis}). Finally, we evaluated the fine-grained level of samples in EPIC-KITCHENS-100-MQA using GPT-4o and showed LLaVAction's robustness across different fine-grained levels (Section \ref{sec:psychometrics_analysis}).
% showing that LLaVAction learns to use the action token to gain action-related visual information.

\vspace{-.1cm}
\subsection{Difficulty levels of fine-grained action understanding analysis}
\label{sec:psychometrics_analysis}

\begin{wraptable}{r}{0.5\linewidth}
\centering
\scriptsize
\vspace{-.3cm}
\begin{tabular}{lcccc}
\toprule
Difficulty levels & 1 & 2 & 3 & 4 \\ 
\midrule
LLaVA-Video Acc. & 0.405 & 0.360 & 0.342 & 0.339 \\
LLaVAction Acc.  & 0.735 & 0.744 & 0.738 & 0.726 \\
\# Samples          & 400   & 3529  & 2836  & 2903  \\
\bottomrule
\end{tabular}
\caption{\textbf{Model performance under different difficulty levels.} LLaVAction is more robust with different semantic similarity between options.}
\vspace{-.3cm}
\label{tab:psychometrics_analysis}
\end{wraptable}

While EPIC-KITCHENS-100-MQA features distinguishing fine-grained actions, individual samples may still vary in difficulty. To understand how baseline models and LLaVAction perform across different difficulty levels, we adopted the concept of psychometric curves~\citep{boring1917chart}. We utilized GPT-4o to rate the difficulty of distinguishing between the options (detailed prompt in Appendix \ref{sec:difficulty_level}) from 1 (very easy) to 4 (hard) and then reported the performances of LLaVA-Video and LLaVAction under different levels on EPIC-KITCHENS-100-MQA (Table \ref{tab:psychometrics_analysis}). LLaVAction not only achieves higher overall accuracy but also maintains more robust performance as difficulty increases, showing less performance degradation compared to baseline models.

\vspace{-.1cm}
\section{Conclusion} 
\vspace{-.1cm}

Recent advances in MLLMs prompted our investigation of their fine-grained action understanding abilities. Through our proposed EPIC-KITCHENS-100-MQA benchmark, which uses similar actions as distractors, we reveal that state-of-the-art MLLMs face significant challenges in action discrimination tasks. We address these limitations by introducing specialized data reformulation strategies and action-aware architectural components that substantially enhance MLLM action recognition capabilities. The resulting LLaVAction model achieves robust performance and demonstrates strong generalization across our benchmark, three additional action recognition datasets, and ten comprehensive MLLM video understanding benchmarks.

%The development of MLLMs motivated us to explore their fine-grained action understanding ability. We show that when we sample similar answers as distractors in our proposed EPIC-KITCHENS-100-MQA benchmark, leading MLLMs struggle to identify the correct actions. We then designed a new MLLM data reformulation and action-related model designs that greatly improved the ability to perform action recognition with MLLMs. The resulting model LLaVAction showed strong performance and generalization ability on EPIC-KITCHENS-100-MQA, 3 action recognition benchmarks and 10 MLLM video understanding benchmarks.

%Specifically, we obtained SOTA performance on the proposed EPIC-KITCHENS-100-MQA with 21 points in accuracy over GPT-4o. We achieved SOTA performances on three action recognition benchmarks to show LLaVAction's superiority and generalization ability in comparison to specialized models. Furthermore, we showed consistent improvements over the LLaVA-Video baseline on 10 zero-shot benchmarks covering detailed caption, open-ended question answering and multi-choice question answering tasks, which further validates the effectiveness of LLaVAction as well as our training paradigm.

\section{Ethics statement}

This work utilizes established human activity benchmarks and trains the resulting LLaVAction model exclusively on data that contain common daily activities. Given the benign nature of the training data and the focus on routine human behaviors, we anticipate minimal ethical concerns. All human activity data used come from publicly available datasets that have previously been used by the research community.

% \section{Reproducibility statement}

% To ensure reproducibility, we provide an anonymized codebase as supplementary material containing the core components of this work, including the LLaVAction model architecture and benchmark construction methodology. Complete code, trained models, and detailed experimental configurations will be made publicly available upon paper acceptance.

\section*{Acknowledgments}
We thank the Swiss AI Initiative Project ID a03 and a144 from the Swiss National Supercomputing Centre (CSCS); H.Q. and A.M. thank the Boehringer Ingelheim Fonds PhD stipend; M.W.M. thanks the Vallee Foundation; M.W.M. and A.M. thank the SNSF by grant No. 320030-227871.

\section*{Author Contributions:} Conceptualization: S.Y., H.Q., A.M., M.W.M.; Methodology \& Software: H.Q., S.Y., A.M., M.W.M.; Experiments: H.Q., S.Y.; Writing: H.Q., S.Y., M.W.M, A.M.; Visualization: H.Q., S.Y., M.W.M.; Funding acquisition: M.W.M., A.M.

\bibliography{iclr2026_conference}
\bibliographystyle{iclr2026_conference}

\appendix
\newpage
\input{appendix}

\end{document}

%% file: math_commands.tex
\usepackage{amsmath,amsfonts,bm}

\def\eqref#1{equation~\ref{#1}}
\def\1{\bm{1}}

\DeclareMathAlphabet{\mathsfit}{\encodingdefault}{\sfdefault}{m}{sl}
\SetMathAlphabet{\mathsfit}{bold}{\encodingdefault}{\sfdefault}{bx}{n}

%% file: appendix.tex
\setcounter{page}{1}
\label{sec:appendix_section}

\section{Methods: Licensing information}
\label{sec:licensing}

\begin{table}[h] 
\centering 
\begin{tabular}{ll} \toprule \textbf{Code/Dataset} & \textbf{License} \\
 \midrule
 \href{https://github.com/LLaVA-VL/LLaVA-NeXT/tree/79ef45a6d8b89b92d7a8525f077c3a3a9894a87d}{LLaVA-NeXT} & \href{https://www.apache.org/licenses/LICENSE-2.0}{Apache-2.0} \\ 
\href{https://github.com/zhaoyue-zephyrus/AVION}{AVION} & \href{https://opensource.org/license/mit}{MIT}  \\ 
\href{https://github.com/JacobChalk/TIM}{TIM} &  \href{https://creativecommons.org/licenses/by-nc-sa/4.0/deed.en}{CC-NC-SA-4.0} \\ 
EPIC-KITCHENS-100 \\ \href{https://data.bris.ac.uk/data/dataset/3h91syskeag572hl6tvuovwv4d}{Dataset 55} and \href{https://data.bris.ac.uk/data/dataset/2g1n6qdydwa9u22shpxqzp0t8m}{extended} &  \href{https://www.nationalarchives.gov.uk/doc/non-commercial-government-licence/version/2/}{NC-Government}  \\ 
EPIC-KITCHENS-100 \\ \href{https://github.com/epic-kitchens/epic-kitchens-100-annotations}{Annotations} & \href{https://creativecommons.org/licenses/by-nc-sa/4.0/deed.en}{CC-NC-SA-4.0} \\ 
\href{https://github.com/EvolvingLMMs-Lab/lmms-eval}{lmms-eval} & \href{https://opensource.org/license/mit}{MIT} and \href{https://www.apache.org/licenses/LICENSE-2.0}{Apache-2.0} \\ 
\bottomrule 
\end{tabular} 
\caption{\textbf{List of codes and datasets with their corresponding licenses.}} 
\label{tab:licenses} 
\end{table}

\section{Methods: GPT-4o distillation.}
\label{sec:gpt_distillation}

Due to the cost, we sample 4 frames per annotated video clip to go over the training set of EPIC-KITCHENS-100.

We first get the caption corresponding to all video clips in the training set, and then we use the captions obtained to create open-ended question-answers. We show the corresponding prompts for the generations of captions and open-ended question-answers as follows.

\subsection{GPT-4o and GPT-4o-mini annotation prompt for the caption task.}

\begingroup
\ttfamily
You are viewing video frames from an egocentric perspective and you are the person. Describe the video frames in detail and reason about the actions you are performing. You will be provided with the human-annotated ground-truth for the action, but you should independently come to your own conclusion.

If you disagree with the human annotation, indicate "true" in the "disagree\_with\_human\_annotation" field of your response, and provide your reasoning without mentioning the ground-truth answer. This will keep your reasoning clean. If you agree with the human annotation, indicate "false" in the "disagree\_with\_human\_annotation" field and provide your reasoning without referencing the ground-truth to maintain a clean description. The true ground-truth action is \{gt\_answer\}.
Your reasoning steps should include supporting evidence for the action, such as the duration of the video, the sequence of actions the person performs, the objects they interact with, and the overall context of the video.

As a general guideline, for videos longer than 3 seconds, provide detailed reasoning steps, and for videos shorter than 3 seconds, generate less detailed reasoning.
The video duration is \{end\_second - start\_second:.3f\} seconds.
Make sure you use the first-person perspective in your reasoning.
\endgroup

\subsection{GPT-4o and GPT-4o-mini annotation prompt for Open-ended question answering}

\begingroup
\ttfamily
Your job is to create 3 question-answer pairs based on the text below. The text contains a first-person narrative of video frames from an egocentric perspective of a person interacting with objects in a kitchen.
{caption\_text}
You can ask questions such as:
What object am I interacting with?
What objects are visible in the video?
What is the sequence of the atomic actions I am performing? Make sure your questions can be answered based on the information provided in the text. Do not ask questions that require additional context or information beyond what is given.
\endgroup

\section{Methods: LLaVAction task prompts}

\subsection{LLaVAction caption prompt}
\label{sec:llavaction_caption_prompt}

\begingroup
\ttfamily
Describe in details what you see from the video frames. Try to focus on what you are doing.
\endgroup

\subsection{LLaVAction perspective prompt}
\label{sec:perspec_prompt}

\textbf{Egocentric.} For the EgoSchema benchmark, given that our LLaVAction-7B is trained with egocentric perspective prompt on EPIC-KITCHENS-100, we use the same egocentric perspective prompt when we evaluate our model on EgoSchema benchmark.

\begingroup
\ttfamily
You are seeing this video from egocentric view and you are the person. Your hands are sometimes interacting with objects. What action are you doing?
\endgroup

\textbf{Allocentric.}

\begingroup
\ttfamily
The video is taken from egocentric view. The person's hands are sometimes interacting with objects. What action is the person doing?
\endgroup

\subsection{LLaVAction temporal detection prompt}
\label{sec:temporal_detection}

\begingroup
\ttfamily
The provided video contains an action \{ACTION NAME\} that lasts 2.96 seconds. What is the relative start and end time of the action in seconds? Format it as 'start\_timestamp: end\_timestamp' and round to 2 decimal places.
\endgroup

\subsection{LLaVAction direct prediction prompt}
\label{sec:direct_prediction}

\begingroup
\ttfamily
What action are you performing? Give a short sentence such as 'move knife'.
\endgroup

\subsection{LLaVAction prior action learning prompt}
\label{sec:prior_action}

\begingroup
\ttfamily

\{prev2\_offset\} seconds ago, you started an action \{prev2\_narration\}. \{prev1\_offset\} seconds ago, you started an action \{prev1\_narration\}. What action are you currently performing? Here are the options of actions you can select:
\endgroup

\subsection{Difficulty level assessment prompt}
\label{sec:difficulty_level}

\begingroup
\ttfamily
You are analyzing a multiple-choice question for fine-grained action recognition. Your task is to rate the difficulty of distinguishing between the options based on how similar they are to each other and the ground truth answer.

Ground Truth Answer: gt\_answer

Options:
options\_text

Please analyze the semantic similarity between the options and rate the difficulty on a scale of 1-4:

- 1 (Very Easy): Options are very different from each other and the correct answer is obvious

- 2 (Easy): Options have clear differences, correct answer is fairly obvious  

- 3 (Medium): Options have moderate similarity, requires some careful consideration

- 4 (Hard): Options are quite similar, subtle differences make it challenging

Consider factors like:

- Semantic similarity between action descriptions

- Specificity vs generality of actions

- Whether options describe similar but distinct actions

- How confusable the distractors are with the ground truth

Respond with only a single number (1-4) representing the difficulty score.
\endgroup

\section{AVION as OOD distractors}
\label{sec:OOD}

We utilized TIM's~\citep{chalk2024tim} predictions to build our EPIC-KITCHENS-100-MQA benchmark. When we use similar ideas to build the hard distractor training set, it results in IID setting if we still use TIM's~\citep{chalk2024tim} predictions. {\color{revision}{Methods could directly overfit on TIM's choice combination to obtain better performances instead of contrasting fine-grained actions. Therefore, we used AVION's~\citep{zhao2023training} predictions during training to serve as an OOD setting. 

Here, we provide three experiments to support using AVION can be an OOD setting. First, we calculated the top-1 agreement percentage (65\%) and top-5 overlap percentage (45\%)  between AVION and TIM in EPIC-KITCHENS-100’s validation set, suggesting a considerable difference in distribution, especially when K is larger. 

Additionally, we computed the Jensen–Shannon Divergence (JSD) between the softmax outputs of Avion and Tim across the validation set (9668 samples). The mean JSD was $0.674 \pm 0.089$, with a 95\% confidence interval of [0.672, 0.765]. We obtained a p-value $< 0.001$ and a Cohen’s $d$ of 7.57, indicating a large and statistically significant difference between Avion and Tim. As a consequence, their generated distractors should be seen as coming from two different distributions.

An important signature of the IID vs. OOD argument is that OOD is less vulnerable to overfitting when giving the model more chances to explore the training data. Following our OOD setting that uses Avion distractors for training and TIM distractors for testing, we performed experiments that vary K in both training and testing, for K = 5, 10, 20, we got 74.3, 69.5, 64.1, respectively. Since test-time distractors are generated by TIM and training-time distractors are generated by AVION, we believe increasing K in training introduces overfitting, and thus it does not generalize well to TIM’s distractors, which further supports our rationale for the OOD setting.

}}

\section{Additional ablations}

{\color{revision}{
\subsection{Action token design ablation}

Our action token design (Section~\ref{sec:our_model}) effectively sees and encodes action-related video information, and hence benefits the question-answering task. To further support our action token design, we implement several other token-aggregation methods. Specifically, our action token is one learnable token added between the visual tokens and text tokens and is supervised in the last layer with the action classification loss. We hence implement three variants: 1) adding three action tokens and supervising them with verb, noun, and action separately in the last layer; 2) adding one action token and supervising it in the first layer of the MLLM; adding one action token and supervising it across all MLLM layers. The results are in Table \ref{tab:action_token_ablation} on our benchmark. We can see our action design performs the best while keeping simplicity. }}

\begin{table}[h]
\centering
{\color{revision}{
\begin{tabular}{ll}
\toprule
Token designs & Acc. \\ 
\midrule
3 tokens, last layer                     & 	68.8 \\%/73.4    \\
1 token, first layer      & 	66.2  \\%/68.2                  \\
1 token, all layers     & 31.7 \\%/73.1              \\
1 token, last layer (Ours)     & \textbf{69.1} \\%/72.1               \\
\bottomrule
\end{tabular}
\caption{\textbf{Action token design ablation study}. The top 1 percent accuracy is shown.
}
\label{tab:action_token_ablation}
}}
\end{table}

\subsection{Leave-one-out Ablation}
\label{sec:leave_one_out}

Remarkably, the 10-point gain over our baseline model cannot be attributed to only a single factor. We took our full model, i.e., the base plus all added methods, which we call \textbf{LLaVAction-7B}, and performed a leave-on-out ablation (Table~\ref{tab:leave_one_ablation}). Given our additions adds negligible overhead in the inference time (only one special vision token added to the baseline model), we then suggest using our full LLaVAction-7B and techniques in downstream tasks.

\begin{table}[h]
\centering
\begin{tabular}{ll}
\toprule
                               LLaVA-Video-7B & Acc. \\ 
\midrule
Full (LLaVAction-7B)                       & 74.1 \\%/73.4    \\
Full w/o adversarial distractors (AVION)      & 69.7  \\%/68.2                  \\
Full w/o action token     & 73.6 \\%/73.1              \\
Full w/o temporal detection     & 72.2 \\%/72.1               \\
Full w/o GPT-4o-based reformulation     & 73.2 \\%/73.2               \\
Full w/o direct prediction   & 73.2 \\%/72.8               \\
Full w/o temporal order learning & 72.3 \\%/ n/a              \\
\bottomrule
\end{tabular}
\caption{\textbf{Leave-one-out ablation study}. Full denotes having all the proposed methods. In each row we drop one method from the full method and report the resulting performance. 16 frames were used for both training and testing, and the percent accuracy is shown. %The value after the \ is the test-time augmentation performance.
}
\label{tab:leave_one_ablation}
\end{table}

{\color{revision}{
\subsection{Structured action prediction ablation}

Our proposed two-stage pipeline enables MLLMs to fairly compare and outperform other SOTA action recognition methods. MLLMs directly output free texts, which makes it hard to find an exact match with the action labels in the dataset, especially when the action space is huge and fine-grained. With an external model applied in the first stage to filter out easy, irrelevant actions, MLLMs can mainly focus on differentiating between the hard distracting actions. To support that, we evaluated LLaVA-Video-7B with the same external model on the EPIC-KITCHEN action recognition benchmark (Table \ref{tab:structured_action_ablation}). We can see that the performance of LLaVA-Video-7B is much worse, even with an external model, showing that it struggles to solve the hard distractors. Meanwhile, we further implement another way (denoted as ‘Multi-round appending’) to achieve structured action output. Specifically, we first prefill and store the KV Cache for video+prompt+question to avoid repeated computation. After that, we append each action class to compute the text cross-entropy loss. The action class with the lowest loss is selected as the final action prediction. We test for both the zero-shot LLaVA-Video-7B and our LLaVAction models on the EPIC-KITCHEN action recognition benchmark (Table \ref{tab:structured_action_ablation}). The results show our two-stage action prediction pipeline can obtain much better performance under both fine-tuned and zero-shot settings. Most importantly, the multi-round appending is extremely time-consuming. Although KV Cache storage avoids computing video+prompt+question repeatedly, the model still needs to infer 4K times to obtain the correct answer. Evaluating the model on EPIC-KITCHENS-100’s validation set (9668 samples) takes around 820 GPU hours when using the multi-round appending approach. In comparison, our two-stage approach only takes 4.3 GPU hours, making our two-stage method 190 times faster.

}}

\begin{table}[h]
\centering
{\color{revision}{
\begin{tabular}{ll}
\toprule
Methods & Acc. \\ 
\midrule
\multicolumn{2}{c}{\textit{LLaVA-Video-7B}} \\ \midrule
Multi-round appending                     & 	10.3 \\%/73.4    \\
Two-stage (Ours)      & 	26.5  \\ \midrule
\multicolumn{2}{c}{\textit{LLaVAction-7B}} \\ \midrule
Multi-round appending     & 40.0 \\%/73.1              \\
Two-stage (Ours)     & \textbf{58.3} \\%/72.1               \\
\bottomrule
\end{tabular}
\caption{\textbf{Structured action prediction ablation study}. The top 1 percent accuracy is shown.
}
\label{tab:structured_action_ablation}
}}
\end{table}

\begin{table*}[ht]
\centering
\begin{tabular}{lcc}
\toprule
Methods & Action label & Action narration \\
\midrule

zero-shot LLaVA-Video-7B & 26.5  & 35.7      \\
zero-shot LLaVA-OV-7B    & 19.6  & 28.9           \\
zero-shot LLaVA-OV-0.5B  & 24.8  & 32.0       \\ 

\bottomrule
\end{tabular}
\caption{\textbf{Quantitative results for action label vs. action narration.} Models are inferred with eight frames as inputs.}
\label{tab:key_narration_comparison}
\end{table*}

\section{Qualitative examples}
\label{sec:qualitative_examples}

\subsection{Comparing action narration and action label in EPIC-KITCHENS-100}
\label{sec:action_label}

The action labels in EPIC-KITCHENS-100 originate from the raw action narrations that are curated and compressed by a combination of word clustering and iterative manual refinement~\citep{damen2022rescaling}. However, this compression might change the semantic meaning of both nouns, verbs and the way they are combined. As a result, large language models that are sensitive to the meaning of words can be misled (see comparisons in Appendix Figure~\ref{fig:narration_official_key} and Table \ref{tab:key_narration_comparison}). While we show SOTA results using the action label, we note that we can achieve better performance if we use the uncompressed, original narrations. We hope that our work could inspire future work to study the best text representation of actions to train and evaluate MLLMs in action recognition.

Qualitatively: we illustrate some examples of choices represented in the action label manner (Appendix Figure \ref{fig:narration_official_key}). We show the ground truth option in blue and the prediction of LLaVAction-7B in pink. We can see LLaVAction-7B's predictions also make sense in those examples and hence cause ambiguity across choices. Instead, the corresponding action narration fits better to the language's nature and can better describe the video content with less ambiguity. 

Quantitatively: furthermore, we also quantify MLLMs' zero-shot performance (LLaVA-OV-0.5B, LLaVA-OV-7B, LLaVA-Video-7B) when using action labels or action narrations as inputs (Table \ref{tab:key_narration_comparison}). The inferior zero-shot performance of all 3 evaluated models when tested on the action labels as action representation supports our qualitative observations that action labels are less ideal than narrations for MLLMs.

\subsection{Different choices comparison}

\begin{figure*}[h!]
    \centering
    \includegraphics[width=\linewidth]{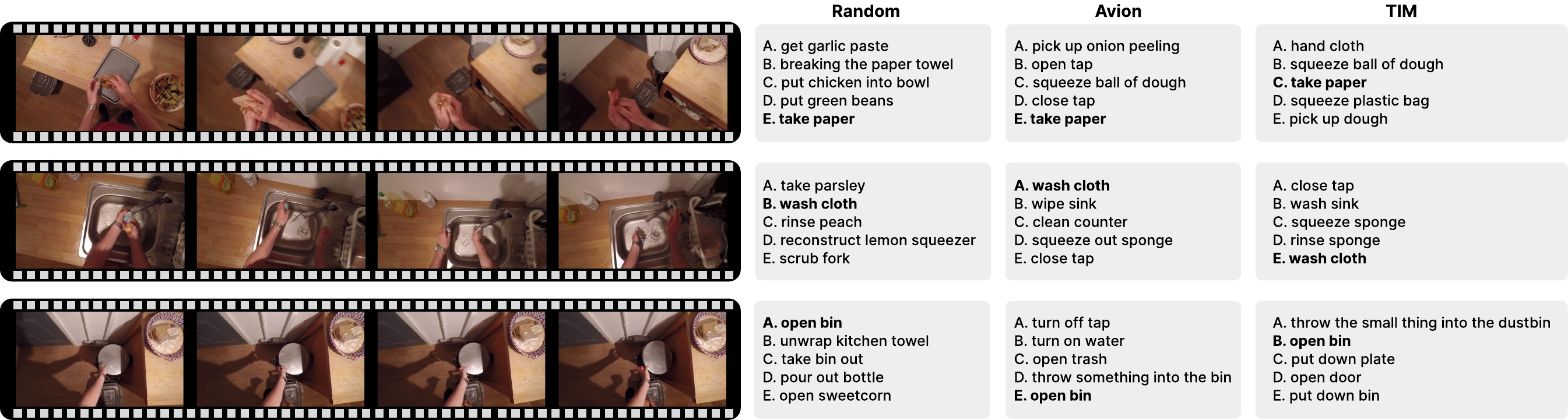}
    \caption{\textbf{TIM's choices are harder than AVION and random} by introducing more visually similar objects and actions. Bold option denotes ground truth.}
    \label{fig:choice_comparison}
\end{figure*}

Here, we show examples of choices generated by random sampling, AVION top-5 predictions, and TIM top-5 predictions (Appendix Figure \ref{fig:choice_comparison}). We can see that the randomly selected choices have many trivial choices that can be easily distinguished with the correct answer. In comparison, choices generated based on AVION and TIM top-5 predictions become much more similar to the correct answer and exhibit features such as similar object/scene, temporal orders or object relationships that are emphasized by other benchmarks.

\subsection{LLaVAction Caption}
\begin{figure*}[h!]
    \centering
    \includegraphics[width=\linewidth]{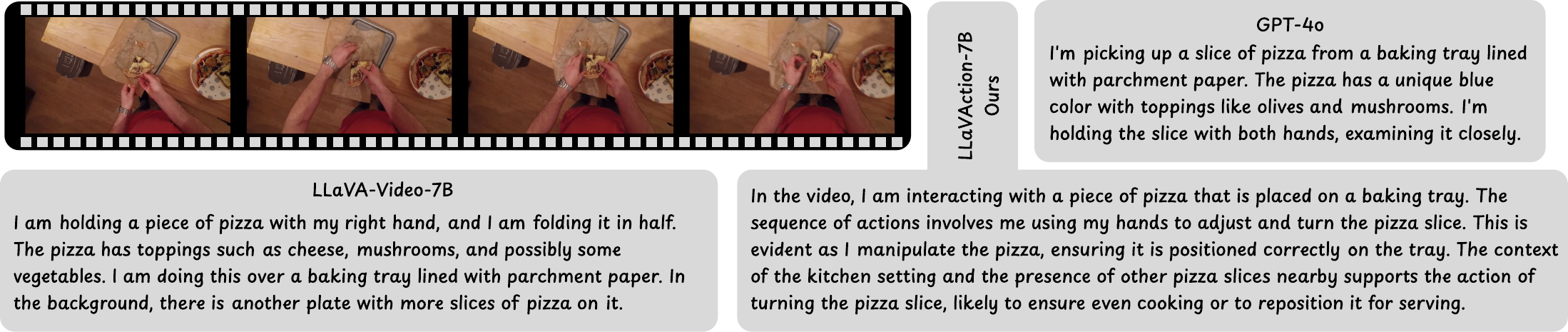}
    \caption{\textbf{The captioning capability can provide insights into models.} A comparison among LLaVAction-7B, GPT-4o and LLaVA-Video-7B.}
    \label{fig:llavaction_caption_task}
\end{figure*}

Here, we show one video caption example of different models including GPT-4o, LLaVA-Video-7B and our LLaVAction-7B (Appendix Figure \ref{fig:llavaction_caption_task}). We can see the interacting object (pizza piece) is pretty small in the video and there are also many other distracting objects. Both GPT-4o and LLaVA-Video-7B cause 'hallucinations' in their descriptions. For example, GPT-4o thinks the person holds the slice with both hands. Instead, LLaVAction-7B still retains the video caption ability and can generate plausible descriptions of the video.

\section{Sub-category performance comparisons on the additional benchmarks}
\label{sec:additional_benchmarks_2}

Snice MVBench and LongVideoBench also have sub-category measurements with many of them related to action understanding, we also show the sub-category performances on these two benchmarks in this section.

\subsection{Performance comparison on sub-categories of MVBench}

Here we show the performance comparison between LLaVA-Video-7B and our LLaVAction-7B on sub-categories of MVBench. We can see LLaVAction-7B boost the performance on many action-related categories such as action count, action sequence and fine-grained action, etc.

\begin{table*}[ht]
\centering
\begin{tabular}{lccc}
\toprule
Tasks        & LLaVA-Video-7B & LLaVAction-7B (Ours) & Difference \\
\midrule
Action antonym           & 76.0     & 75.0   & -1.0   \\
Action count             & 57.0     & 65.0   & 8.0   \\
Action localization      & 61.0     & 63.5   & 2.5   \\
Action prediction        & 62.0     & 59.0   & -3.0      \\
Action sequence          & 70.5     & 72.5   & 2.0     \\ 
Character order          & 74.5     & 79.0   & 4.5 \\
Counterfactual inference & 50.0     & 52.0   & 2.0     \\
Egocentric navigation    & 30.5     & 28.0   & -2.5  \\
Episodic reasoning       & 53.5     & 54.0   & 0.5   \\
Fine-grained action      & 48.0     & 49.0   & 1.0   \\
Fine-grained pose        & 54.5     & 61.5   & 7.0     \\
Moving attribute         & 71.0     & 72.5   & 1.5      \\
Moving count             & 44.0     & 43.0   & -1.0     \\ 
Moving direction         & 35.5     & 31.0   & -4.5    \\
Object existence         & 60.0     & 59.0   & -1.0     \\
Object interaction       & 84.5     & 83.5   & -1.0     \\
Object shuffle           & 41.5     & 44.0   & 2.5     \\ 
Scene transition         & 93.5     & 90.5   & -3.0     \\
State change             & 54.0     & 61.5   & 7.5     \\
Unexpected action        & 81.5     & 79.0   & -2.5     \\
\bottomrule
\end{tabular}
\caption{\textbf{Sub-category comparison with LLaVA-Video-7B on MVBench.}}
\label{tab:mvbench}
\end{table*}

\subsection{Performance comparison on sub-categories of LongVideoBench}

Here we show the performance comparison between LLaVA-Video-7B and our LLaVAction-7B on sub-categories of LongVideoBench. We can see LLaVAction-7B also boosts the performance on many action-related categories such as event before/after, text-referred object attribute, and object-before/after object.

\begin{table*}[ht]
\centering
\begin{tabular}{lccc}
\toprule
Tasks        & LLaVA-Video-7B & LLaVAction-7B (Ours) & Difference \\
\midrule
Event-Referred object                    & 72.31  & 69.23  & -3.08         \\
Event-Before/after event                 & 67.02  & 67.02  & 0.0        \\
Object-Referred event                    & 67.82  & 64.37  & -3.45        \\
Object-Before/after object               & 57.58  & 59.09  & 1.52            \\
Scene-Referred object attribute          & 71.59  & 70.46  & -1.14          \\ 
Scene-Referred event                     & 72.04  & 66.67  & -5.38      \\
Scene-Referred object                    & 63.89  & 63.89  & 0.0          \\
Scene-Referred object attribute change   & 55.56  & 52.78  & -2.78        \\
Scene-Referred object tracking           & 65.43  & 66.67  & 1.23        \\
Sequence of scenes                       & 41.24  & 41.24  & 0.0        \\
Text-Referred object attribute           & 59.49  & 62.02  & 2.53          \\
Text-Referred event                      & 56.92  & 56.92  & 0.0           \\
Text-Referred object                     & 59.21  & 59.21  & 0.0           \\ 
Event before/after text                  & 50.68  & 58.90  & 8.22          \\
Object before/after text                 & 58.11  & 52.70  & -5.41           \\
Text-Referred object attribute change    & 47.56  & 50.00  & 2.44           \\
Text-Referred object tracking            & 32.88  & 32.88  & 0.0          \\ 
\bottomrule
\end{tabular}
\caption{\textbf{Sub-category comparison with LLaVA-Video-7B on LongVideoBench.}}
\label{tab:longvideobench}
\end{table*}

\subsection{Performance comparison on sub-categories of VideoMME}

Here we show the performance comparison between LLaVA-Video-7B and our LLaVAction-7B on sub-categories of VideoMME. Our model did not improve the action recognition performance on VideoMME possibly due to the domain gap between VideoMME and EPIC-KITCHENS-100.

\section{Extended Discussion}

\textbf{Egocentric vs. Allocentric perspective.} MLLM can be assigned with different roles before seeing the video. Since the videos are taken from the first-person perspective, we believed the egocentric perspective aligns better with the LLM pretraining data. Therefore, we switch from the third-person (allocentric) perspective to the first-person (egocentric) perspective to better guide the model. We present the prompts in the Appendix~\ref{sec:perspec_prompt}. When we fix the distractors from random sampling, using the egocentric prompt gives a 0.5 point improvement over using the allocentric prompt on EPIC-KITCHENS-100-MQA.

\textbf{Alternative approaches we tested.}
\label{disc:alt}
We also tested a few alternative approaches to improve MLLMs in our benchmark. We tried self-consistency predictions, which do not yield improvements, perhaps due to the task being vision-centric. Additionally, we explored multi-modal chain-of-thought (COT) reasoning by prompting the model to generate a caption prior to addressing the multi-question answering task. However, we found that the model exhibited reluctance to perform this action, despite being capable of generating captions or answering multi-choice questions independently. 
A variant of it is to inference the model twice, so we have the caption first and feed that into the instruction of answering multi-choice question task, similar to~\citep{zhang2024multimodalchainofthoughtreasoninglanguage}. While a minor improvement was observed, we think it is not worth the 2X compute. We believe that video action recognition is a good way to explore video reasoning for MLLMs. However, we leave COT improvements on this task for future work.

%We also attempted to apply DPO~\citep{zhang2024directpreferenceoptimizationvideo} in order to facilitate the model's ability to differentiate between an action and its neighboring actions. However, no significant improvement was observed.

\begin{table*}[t]
\centering
\small
\begin{tabular}{lccc}
\toprule
Tasks        & LLaVA-Video-7B & LLaVAction-7B (Ours) & Difference \\
\midrule
Categories: Artistic Performance & 68.9 & 69.4 & 0.5 \\
Categories: Film \& Television & 71.7 & 72.8 & 1.1 \\
Categories: Knowledge & 76.0 & 75.4 & -0.6 \\
Categories: Life Record & 71.7 & 71.7 & 0.0 \\
Categories: Multilingual & 67.8 & 61.1 & -6.7 \\
Categories: Sports Competition & 65.6 & 66.4 & 0.8 \\
Task Categories: Action Reasoning & 69.5 & 69.1 & -0.4 \\
Task Categories: Action Recognition & 69.0 & 68.4 & -0.6 \\
Task Categories: Attribute Perception & 83.8 & 83.3 & -0.5 \\
Task Categories: Counting Problem & 48.1 & 46.3 & -1.8 \\
Task Categories: Information Synopsis & 87.0 & 86.7 & -0.3 \\
Task Categories: OCR Problems & 73.4 & 74.1 & 0.7 \\
Task Categories: Object Reasoning & 73.6 & 73.3 & -0.3 \\
Task Categories: Object Recognition & 75.7 & 77.1 & 1.4 \\
Task Categories: Spatial Perception & 68.5 & 72.2 & 3.7 \\
Task Categories: Spatial Reasoning & 82.1 & 82.1 & 0.0 \\
Task Categories: Temporal Perception & 76.4 & 78.2 & 1.8 \\
Task Categories: Temporal Reasoning & 51.4 & 52.0 & 0.6 \\
Video Sub Categories: Acrobatics & 65.6 & 64.4 & -1.2 \\
Video Sub Categories: Animation & 58.9 & 60.0 & 1.1 \\
Video Sub Categories: Astronomy & 77.8 & 77.8 & 0.0 \\
Video Sub Categories: Athletics & 66.7 & 73.3 & 6.6 \\
Video Sub Categories: Basketball & 54.4 & 51.1 & -3.3 \\
Video Sub Categories: Biology \& Medicine & 78.9 & 78.9 & 0.0 \\
Video Sub Categories: Daily Life & 78.9 & 75.6 & -3.3 \\
Video Sub Categories: Documentary & 74.4 & 76.7 & 2.3 \\
Video Sub Categories: Esports & 62.2 & 60.0 & -2.2 \\
Video Sub Categories: Exercise & 58.9 & 67.8 & 8.9 \\
Video Sub Categories: Fashion & 68.9 & 70.0 & 1.1 \\
Video Sub Categories: Finance \& Commerce & 80.0 & 80.0 & 0.0 \\
Video Sub Categories: Football & 72.2 & 75.6 & 3.4 \\
Video Sub Categories: Geography & 76.7 & 75.6 & -1.1 \\
Video Sub Categories: Handicraft & 77.8 & 76.7 & -1.1 \\
Video Sub Categories: Humanity \& History & 66.7 & 67.8 & 1.1 \\
Video Sub Categories: Law & 82.2 & 80.0 & -2.2 \\
Video Sub Categories: Life Tip & 70.0 & 67.8 & -2.2 \\
Video Sub Categories: Literature \& Art & 80.0 & 73.3 & -6.7 \\
Video Sub Categories: Magic Show & 62.2 & 65.6 & 3.4 \\
Video Sub Categories: Movie \& TV Show & 68.9 & 70.0 & 1.1 \\
Video Sub Categories: Multilingual & 67.8 & 61.1 & -6.7 \\
Video Sub Categories: News Report & 84.4 & 84.4 & 0.0 \\
Video Sub Categories: Other Sports & 72.2 & 72.2 & 0.0 \\
Video Sub Categories: Pet \& Animal & 78.9 & 78.9 & 0.0 \\
Video Sub Categories: Stage Play & 82.2 & 80.0 & -2.2 \\
Video Sub Categories: Technology & 72.2 & 77.8 & 5.6 \\
Video Sub Categories: Travel & 75.6 & 77.8 & 2.2 \\
Video Sub Categories: Variety Show & 65.6 & 67.8 & 2.2 \\
\bottomrule
\end{tabular}
\caption{\textbf{Sub-category comparison with LLaVA-Video-7B on VideoMME.}}
\label{tab:videomme}
\end{table*}

\clearpage
\newpage

\begin{figure*}[ht!]
    \centering
    \includegraphics[width=\linewidth]{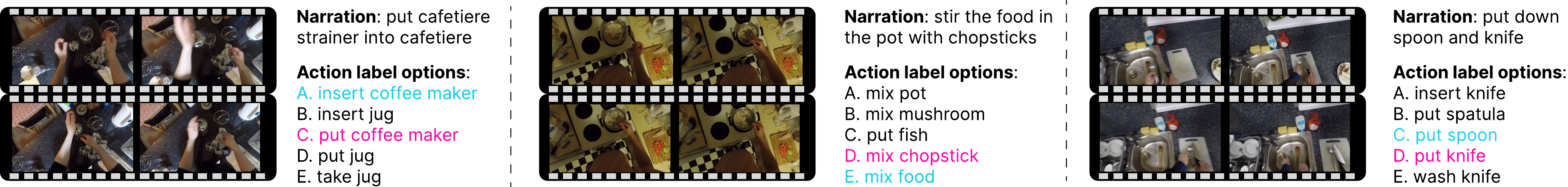}
    \caption{\textbf{Action labels vs. narrations.} Blue option denotes ground truth and the pink option denotes LLaVAction-7B's prediction. Action labels usually reduce multiple nouns into one noun, resulting in ambiguity that could mislead a MLLM. Note that the narration also contains crucial particles with the phrasal verbs to clarify the meaning such as ``put down'', ``put into''.}
    \label{fig:narration_official_key}
\end{figure*}

\begin{figure*}[h!]
    \centering
    \includegraphics[width=.71\linewidth]{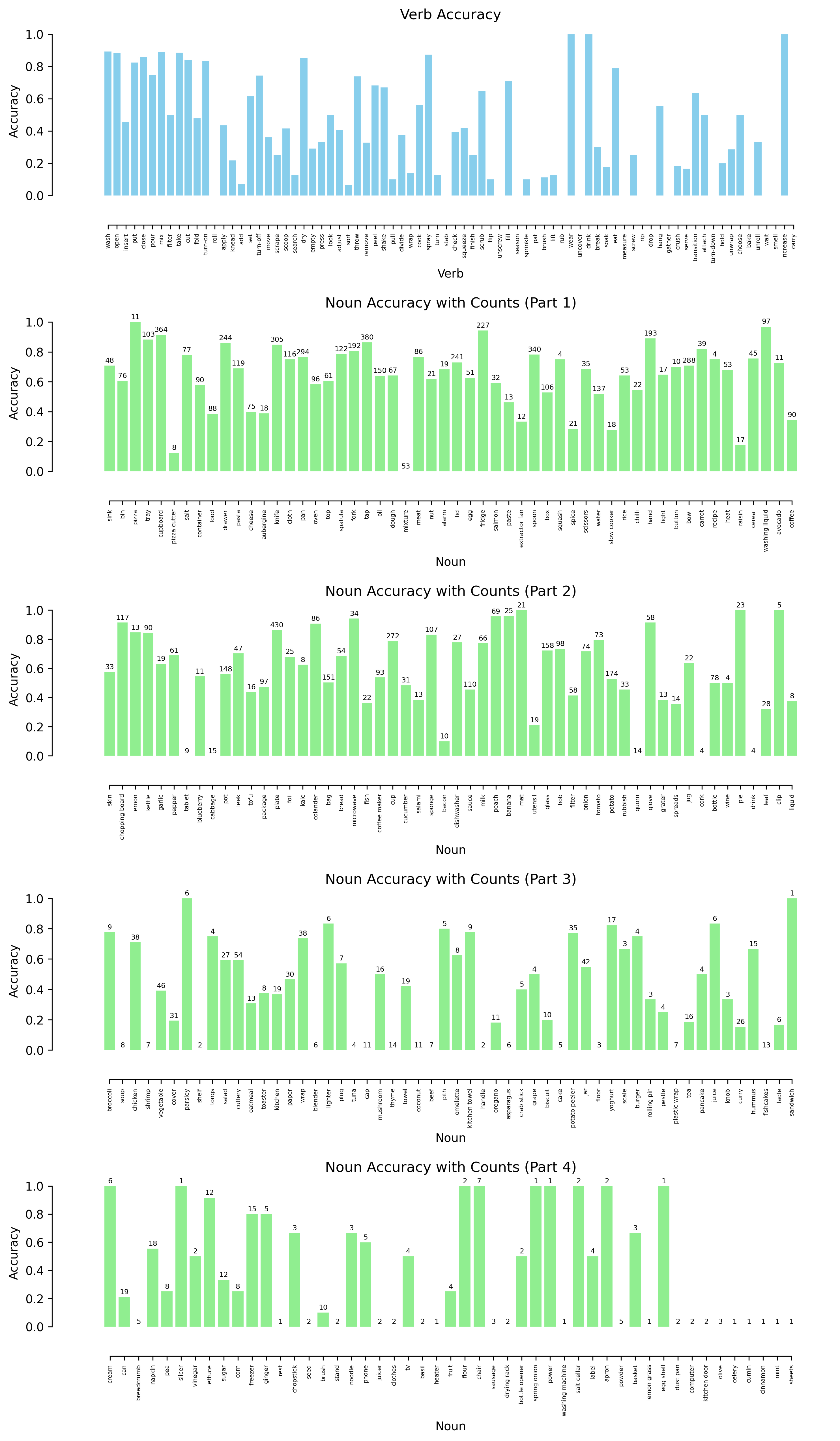}
    \caption{\textbf{Breakdown of the performance of LLaVAction-7B on Verbs and Nouns.} We analyzed the accuracy per verb and noun in EPIC-KITCHENS-100 for that our  LLaVAction-7B (32f), evaluated on the validation set with action labels  (i.e., the model reported in Table~\ref{tab:EPIC-KITCHEN-Original} that achieved 58.3 accuracy). There are more nouns than verbs, thus nouns are shown across four subplots for visualization but otherwise are not separated in an intentional way. The number above each bar is the total per class.}
    \label{fig:llavaction_analysis}
\end{figure*}

\clearpage
\newpage